\documentclass[sn-mathphys,Numbered]{sn-jnl}

\usepackage{graphicx}%
\usepackage{multirow}%
\usepackage{amsmath,amssymb,amsfonts}%
\usepackage{amsthm}%
\usepackage{mathrsfs}%
\usepackage[title]{appendix}%
\usepackage[table]{xcolor}
\usepackage{colortbl}
\usepackage{textcomp}%
\usepackage{manyfoot}%
\usepackage{booktabs}%
\usepackage{algorithm}%
\usepackage{algorithmicx}%
\usepackage{algpseudocode}%
\usepackage{listings}%
\usepackage{subfigure}%
\usepackage{indentfirst}%
\usepackage{tabularx}
\usepackage{adjustbox}

\theoremstyle{thmstyleone}%
\theoremstyle{thmstyletwo}%

\theoremstyle{thmstylethree}%

\definecolor{top1}{RGB}{189,221,189}
\definecolor{top2}{RGB}{216,235,216}
\definecolor{top3}{RGB}{236,245,236}

\newcommand{\hitB}[1]{{\cellcolor{top2}#1}}

\begin{document}

\title[Medical Test-free Disease Detection Based on Big Data]{Medical Test-free Disease Detection Based on Big Data}


\author[a]{\fnm{Haokun} \sur{Zhao}}\email{haokunzhao@link.cuhk.edu.cn}
\equalcont{These authors contributed equally to this work.}
\author[b]{\fnm{Yingzhe} \sur{Bai}}\email{baiyz0107@163.com
}
\equalcont{These authors contributed equally to this work.}
\author[a]{\fnm{Qingyang} \sur{Xu}}\email{123090691@link.cuhk.edu.cn}
\author[a]{\fnm{Lixin} \sur{Zhou}}\email{123090904@link.cuhk.edu.cn}
\author*[b]{\fnm{Jianxin} \sur{Chen}}\email{cjx@bucm.edu.cn}
\author*[a]{\fnm{Jicong} \sur{Fan}}\email{fanjicong@cuhk.edu.cn}

\affil[a]{\orgdiv{School of Data Science}, \orgname{The Chinese University of Hong Kong, Shenzhen}, \orgaddress{\street{2001 Longxiang Road}, \city{Shenzhen}, \postcode{518172}, \state{Guangdong}, \country{China}}}

\affil[b]{\orgdiv{School of Life Sciences}, \orgname{Beijing University of Chinese Medicine}, \orgaddress{\street{11 Beisanhuan East Road}, \city{Beijing}, \postcode{100029}, \country{China}}}


\abstract{Accurate disease detection is of paramount importance for effective medical treatment and patient care. However, the process of disease detection is often associated with extensive medical testing and considerable costs, making it impractical to perform all possible medical tests on a patient to diagnose or predict hundreds or thousands of diseases.
In this work, we propose Collaborative Learning for Disease Detection (CLDD), a novel graph-based deep learning model that formulates disease detection as a collaborative learning task by exploiting associations among diseases and similarities among patients adaptively. CLDD integrates patient-disease interactions and demographic features from electronic health records to detect hundreds or thousands of diseases for every patient, with little to no reliance on the corresponding medical tests.
Extensive experiments on a processed version of the MIMIC-IV dataset comprising 61,191 patients and 2,000 diseases demonstrate that CLDD consistently outperforms representative baselines across multiple metrics, achieving a 6.33\% improvement in recall and 7.63\% improvement in precision. Furthermore, case studies on individual patients illustrate that CLDD can successfully recover masked diseases within its top-ranked predictions, demonstrating both interpretability and reliability in disease prediction. 
By reducing diagnostic costs and improving accessibility, CLDD holds promise for large-scale disease screening and social health security.
}

\maketitle
\noindent Accurate disease detection and diagnosis are crucial for effective medical treatment and patient care. It serves as the foundation upon which appropriate therapeutic strategies are built, enabling healthcare providers to target the root causes of illnesses and deliver precise interventions. Early and accurate diagnosis can significantly improve patient outcomes, reduce the risk of complications, and enhance the overall quality of life for individuals affected by diseases. Moreover, it plays a crucial role in public health efforts by facilitating the identification and management of infectious diseases, thereby preventing their spread within communities.
However, the process of disease detection and diagnosis is often associated with considerable costs \cite{bossuyt2012beyond,qaseem2012appropriate}. Advanced diagnostic technologies, such as imaging studies (e.g., MRI, CT scans) and specialized laboratory tests, require substantial investments in equipment, infrastructure, and highly trained personnel. These expenses are reflected in the fees charged for diagnostic services, which can be a financial burden for patients, especially those without adequate health insurance coverage. Additionally, the time and resources required for thorough diagnostic evaluations can strain healthcare systems, leading to longer wait times and increased administrative costs. 

Given the high cost of disease detection and diagnosis, it is not practical to perform all possible medical tests on a patient to diagnose hundreds or thousands of diseases. In this work, we aim to establish an artificial intelligence tool to detect or predict hundreds or thousands of diseases for a patient by using the existing health data such as electronic health records (EHR), without the need of additional medical tests. EHR has enabled researchers and doctors to build better models for various downstream tasks, such as drug recommendation \cite{yang2021safedrug, shang2019gamenet, mi2024acdnet}, mortality prediction \cite{cai2016real, sottile2021real}, and disease prediction \cite{shen2018utilization, farhan2016predictive}. EHR integrates patient details, reports, and medication records together, leading to large, complex datasets that bring convenience for researchers and doctors. How to utilize patient EHR data effectively, helping doctors reduce the rates of misdiagnosis and missed diagnosis, has been one of the focuses of the healthcare research. Due to the sparse nature of EHR data, many works suffer from data error and limit the potential to make clinical decisions.


\begin{figure}[t]
\centering
\includegraphics[width=\textwidth]{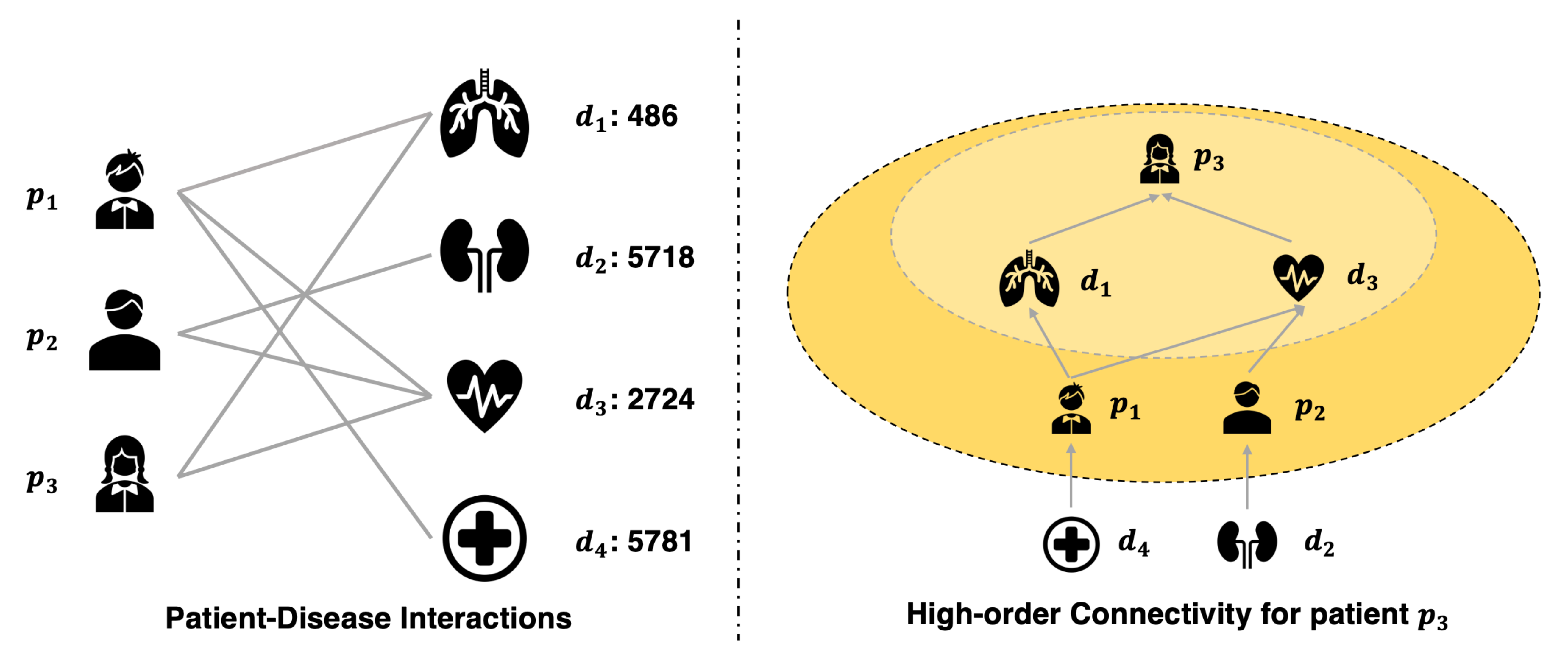}
\caption{\textbf{Illustration of the patient-disease graph and the high-order connectivity for patient $p_3$.} \textbf{Left:} The bipartite patient-disease interaction graph, where each edge denotes a diagnosed relationship. For example, patient $p_3$ is associated with diseases $d_1$ and $d_3$ (ICD-9 codes shown). \textbf{Right:} The tree structure expanded from patient $p_3$. There are two paths connecting $p_3$ and $d_4$ and one path connecting $p_3$ and $d_2$, suggesting patient $p_3$ has a higher inferred likelihood of developing $d_4$ than $d_2$.}\label{high-order}
\end{figure}

In this work, we formulate the disease detection problem as a collaborative learning task \cite{su2009survey,YANG20141} to deeply exploit the knowledge in EHR. The rationale behind this is that different diseases are often correlated and there are often very similar patients. Specifically, many diseases tend to occur together due to shared risk factors, underlying biological mechanisms, or causal relationships. For example, chronic diseases such as diabetes, cardiovascular disease, and kidney disease often coexist in patients \cite{HUANG2023200307}. This co-occurrence is also observed in mental health conditions, where chronic physical illnesses are associated with higher rates of depression and anxiety. Moreover, diseases can be correlated through shared genetic factors and molecular pathways. For instance, mutations in specific genes can lead to multiple phenotypes or increase the risk of developing multiple diseases \cite{sun2014predicting}. Biological networks have been used to model disease associations, revealing that diseases connected through shared genes or pathways often have higher comorbidity rates. On the other hand, in a database, there could be similar patients---they may have similar ages, weights, hobbies, and medical histories. The possibility is higher if the database is larger. Therefore, we exploit the potential associations between hundreds or thousands of diseases and the similarities between thousands or millions of patients.

The idea of adapting collaborative learning has been considered in several previous studies \cite{hao2016comparative,shen2018utilization,sae2022drug}. 
The study of \cite{hao2016comparative} designs different data fusion strategies from heterogeneous resources and integrates them with a collaborative filtering model. The goal is to leverage phenotypic information from electronic medical records and biomedical literature to accelerate the diagnosis of rare diseases. The study of \cite{shen2018utilization} conducts a comparative study between classification and user-based collaborative filtering for clinical prediction. The results suggest that classification-based approaches may be more suitable for clinical prediction tasks where data are incomplete or imbalanced. Despite these attempts, current methods suffer from the following problems: (1) they either form the disease detection problem as a pure user-item collaborative filtering problem without using additional side information or ineffectively encode the side information; (2) the high-order association between diseases and patients cannot be effectively modeled using existing machine learning models, limiting the performance in disease detection. 

Based on the deficiency of previous studies, we proposed our method, collaborative learning for disease detection (CLDD). CLDD constructs the patient-disease relations as a bipartite graph and uses a novel graph neural network to explore the high-order connectivity in the graph, which leads to an effective detection or prediction for potential diseases, avoiding costly medical tests. Fig. \ref{high-order} illustrates the concept of high-order connectivity. The left subfigure is the patient-disease graph extracted from the MIMIC-IV dataset \cite{mimiciv_v2,johnson2023mimic}. Numbers are International Classification of Diseases (ICD) Version 9 for diagnoses. The right subfigure shows the tree structure expanded from $u_3$. The high-order connectivity denotes the path from any node to the target node with a length larger than one. Such high-order connectivity contains rich and meaningful semantics. For example, the path $p_3\leftarrow d_3\leftarrow p_2\leftarrow d_2$ indicates that $p_3$ may have disease $d_2$, since $p_2$ and $p_3$ both have disease $d_3$. Moreover, $p_3$ is more likely to have disease $d_4$ than $d_2$, since there are two paths connecting $<d_4, p_3>$, while only one path connects $<d_2, p_3>$.

We conduct extensive experiments on the MIMIC-IV dataset against several baselines to verify the rationality and effectiveness of our method.

\begin{figure}[t]
\centering
\includegraphics[width=\textwidth]{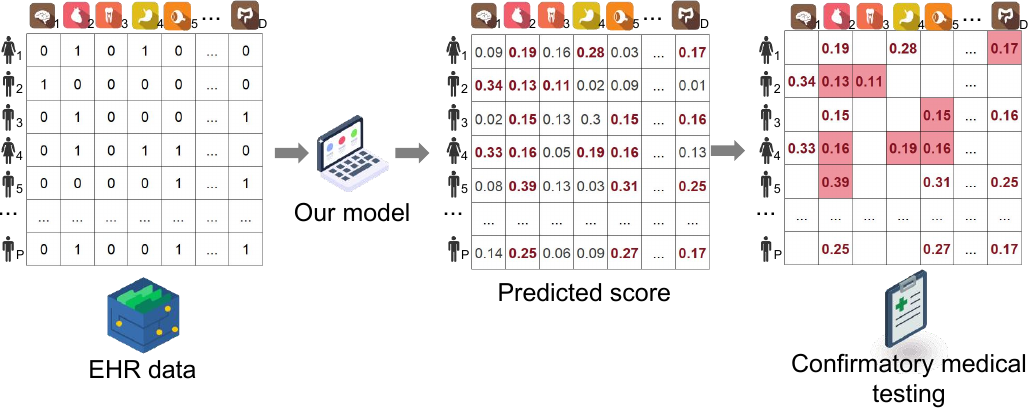}
\caption{\textbf{An illustration of the disease detection workflow.} Starting from a patient-disease interaction matrix extracted from the MIMIC-IV dataset. CLDD outputs a predicted score for every disease-patient pair. The predicted score is ranked for each patient, and the top-$K$ diseases are selected as potential diagnoses for further confirmatory medical testing.}\label{confirmatory medical testing}
\label{fig: 2}
\end{figure}

\section{Results}\label{sec2}
\subsection{Problem and Model Formulation}
\noindent We represent the patient-disease graph as a binary matrix $\boldsymbol{Y}\in\{0,1\}^{P\times D}$, where $P$ and $D$ denote the numbers of patients and diseases, respectively. In the binary matrix $\boldsymbol{Y}$, $y_{pd}=1$ indicates that patient $p$ has been diagnosed with disease $d$, while $y_{pd}=0$ otherwise. 
Note that $y_{pd}=0$ does not mean that the patient does not have the disease. It is possible that the patient did not undergo the corresponding diagnostic tests. We assume that most of the zero-entries in $\boldsymbol{Y}$ correspond to no disease, which is equivalent to assuming a patient does not has the disease with high probability if the corresponding diagnostic test is not conducted. Our method CLDD learns a mapping $\mathcal{M}$ from $\boldsymbol{Y}$ that predicts, for each patient, the likelihood of having any of the $D$ diseases.
As shown in Fig. \ref{fig: 2}, once training is completed, the model $\mathcal{M}$ outputs a ranked list of disease scores for each patient. In practice, patients are advised to undergo confirmatory testing for only the top $K$ predicted diseases with the highest scores (e.g., the top three), thereby significantly reducing both diagnostic cost and examination burden.

\begin{figure}[t]
\centering
\includegraphics[width=\textwidth]{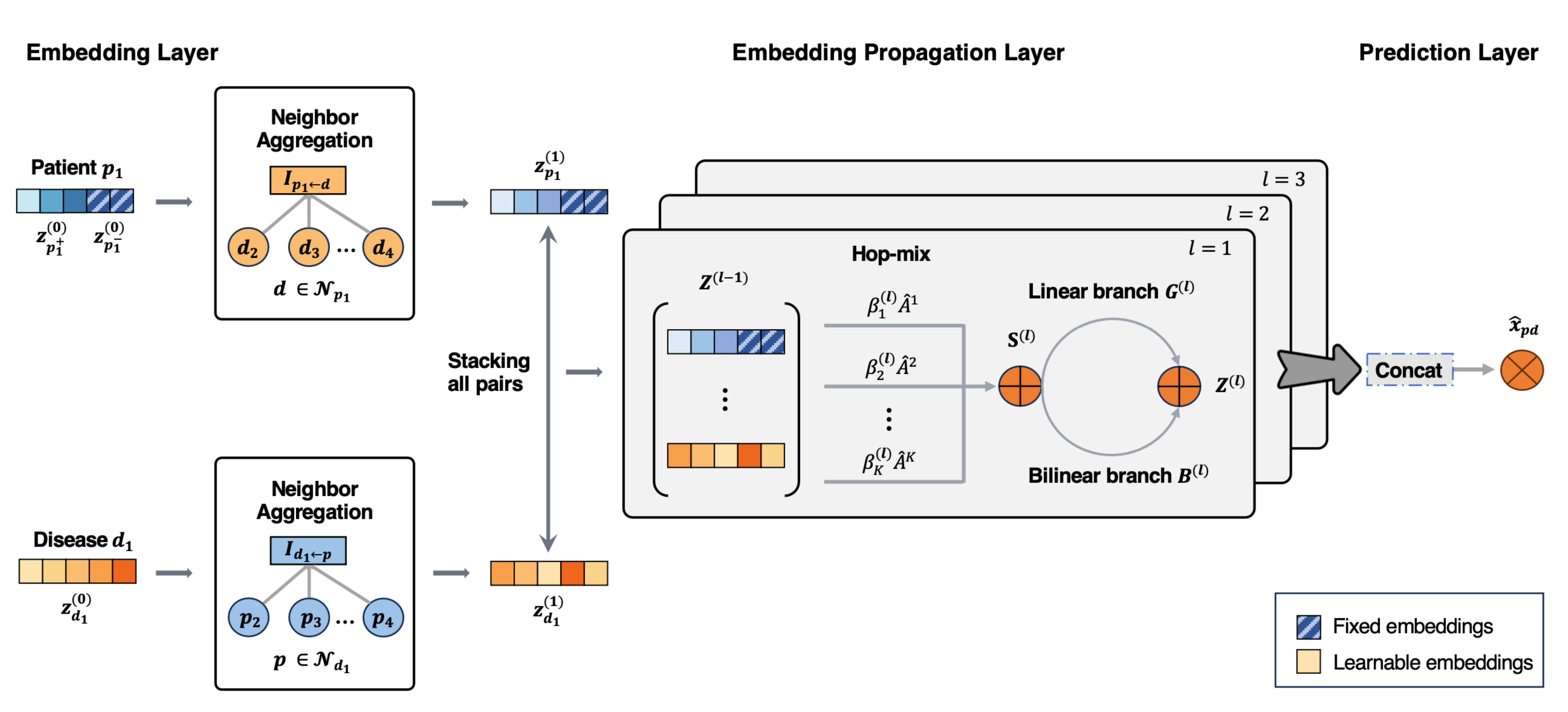}
\caption{\textbf{CLDD model architecture.} The patient embeddings are concatenated by a trainable initialized embedding and fixed attributes. The disease embeddings are all learnable. After the embedding propagation layer, the patient and disease embeddings are updated with a neighbor aggregation layer and refined with multiple embedding propagation layers. In the prediction layer, the outputs are concatenated to make the final prediction.}\label{model architecture}
\end{figure}

Before presenting quantitative results, we briefly summarize the model architecture illustrated in Fig. \ref{model architecture}. CLDD comprises four main components: (1) an embedding layer that initializes patient and disease embeddings; (2) a neighbor aggregation layer that aggregates one-hop relational information; (3) multiple embedding propagation layers that capture higher-order dependencies through iterative propagations; and (4) a prediction layer that concatenate the multi-layer representations to compute the final disease score.

We evaluated CLDD on the MIMIC-IV dataset \cite{mimiciv_v2}, for which we considered $P=61{,}191$ patients and $D=2{,}000$ diseases. The adjacency matrix between patients and diseases is a highly sparse matrix with a sparsity of 0.01093. Quantitative results in Table~\ref{tab1} show that CLDD achieves the best performance among all baselines across five metrics, confirming its ability to recover latent disease relationships from highly sparse interactions.

\subsection{Case Study: Reliable Predictions for Individual Patients}
\noindent To demonstrate the practical reliability of CLDD, we conducted a case study focusing on the model’s ability to recover masked diseases for individual patients. Specifically, for each patient in the MIMIC-IV dataset, we chronologically ordered all disease interaction records based on admission time and assigned the first 80\% of the temporally ordered interactions to the training set, while masking the remaining 20\% for testing. This temporal split ensures that the model learns from historical data to predict future disease occurrences, maintaining temporal consistency and preventing data leakage.

\begin{table}[!htbp]
\centering
\caption{\textbf{Top-5 predicted diseases for the five patients with the highest precision.} Each row corresponds to one patient, and each cell lists the model's top-5 predicted diseases together with their ICD-9/10 codes. Green cells indicate correctly predicted diseases.}
\begin{adjustbox}{max totalheight=\textheight} 
\begin{minipage}{\linewidth}
\begin{tabularx}{\linewidth}{@{}c|*{5}{>{\raggedright\arraybackslash}X}@{}}
\toprule
\textbf{Patient ID} & \textbf{1st} & \textbf{2nd} & \textbf{3rd} & \textbf{4th} & \textbf{5th} \\
\midrule
$20823$ &  Other primary cardiomyopathies (4254) & Acute on chronic systolic heart failure (42823) & Esophageal reflux (53081) & \hitB{Acute on chronic systolic (congestive) heart failure (I5023)} & Diabetes mellitus without mention of complication, type II or unspecified type, not stated as uncontrolled (25000) \\
\midrule
$15$ &  \hitB{Chronic diastolic heart failure (42832)} & \hitB{Do not resuscitate status (V4986)} & Diabetes mellitus without mention of complication, type II or unspecified type, not stated as uncontrolled (25000) & \hitB{Acute on chronic systolic heart failure (42823)} & Abnormal coagulation profile (79092) \\
\midrule
$12991$ &  Diabetes mellitus without mention of complication, type II or unspecified type, not stated as uncontrolled (25000) & \hitB{Diabetes with neurological manifestations, type II or unspecified type, not stated as uncontrolled (25060)} & Old myocardial infarction (412) & Percutaneous transluminal coronary angioplasty status (V4582) & \hitB{Polyneuropathy in diabetes (3572)} \\
\midrule
$15204$ & \hitB{Cirrhosis of liver without mention of alcohol (5715)} & \hitB{Esophageal varices in diseases classified elsewhere, without mention of bleeding (45621)} & Anemia, unspecified (2859) & Iron deficiency anemia, unspecified (2809) & Diabetes mellitus without mention of complication, type II or unspecified type, not stated as uncontrolled (25000) \\
\midrule
$18725$ &  Diabetes mellitus without mention of complication, type II or unspecified type, not stated as uncontrolled (25000) & Background diabetic retinopathy (36201) & Diabetes with ophthalmic manifestations, type II or unspecified type, not stated as uncontrolled (25050) & Cellulitis and abscess of foot, except toes (6827) & \hitB{Atherosclerosis of native arteries of the extremities with ulceration (44023)} \\
\bottomrule
\end{tabularx}
\end{minipage}
\end{adjustbox}
\label{tab:heat_hits}
\end{table}

\begin{table}[!htbp]
\centering
\caption{\textbf{Top-5 predicted diseases for the five patients with the highest recall (restricted to patients with at least five diseases in the test set).} Each row corresponds to one patient, and each cell lists the model's top-5 predicted diseases together with their ICD-9/10 codes. Green cells indicate correctly predicted diseases.}
\begin{adjustbox}{max totalheight=\textheight} 
\begin{minipage}{\linewidth}
\begin{tabularx}{\linewidth}{@{}c|*{5}{>{\raggedright\arraybackslash}X}@{}}
\toprule
\textbf{Patient ID} & \textbf{1st} & \textbf{2nd} & \textbf{3rd} & \textbf{4th} & \textbf{5th} \\
\midrule
$18725$ &  Diabetes mellitus without mention of complication, type II or unspecified type, not stated as uncontrolled (25000) & Background diabetic retinopathy (36201) & Diabetes with ophthalmic manifestations, type II or unspecified type, not stated as uncontrolled (25050) & \hitB{Atherosclerosis of native arteries of the extremities with ulceration (44023)} & Cellulitis and abscess of foot, except toes (6827) \\
\midrule
$30629$ & Other primary cardiomyopathies (4254) & Long-term (current) use of anticoagulants (V5861) & Hypertrophy (benign) of prostate without urinary obstruction and other lower urinary tract symptom (LUTS) (60000) & \hitB{Subendocardial infarction, initial episode of care (41071)} &  \hitB{Other specified forms of chronic ischemic heart disease (4148)} \\
\midrule
$55936$ & Presence of aortocoronary bypass graft (Z951) & Essential (primary) hypertension (I10) & Percutaneous transluminal coronary angioplasty status (V4582) & Long term (current) use of aspirin (Z7982) & \hitB{Long term (current) use of antithrombotics/antiplatelets (Z7902)} \\
\midrule
$44$ &   \hitB{Fever presenting with conditions classified elsewhere (R5081)} & Neutropenia, unspecified (D709) & Unspecified osteoarthritis, unspecified site (M1990) & Headache (R51) & Other medical procedures as the cause of abnormal reaction of the patient, or of later complication, without mention of misadventure at the time of the procedure (Y848) \\
\midrule
$9260$ & Diabetes with neurological manifestations, type II or unspecified type, not stated as uncontrolled (25060) &  \hitB{Unspecified osteomyelitis, ankle and foot (73027)} &  \hitB{Other bone involvement in diseases classified elsewhere (7318)} &  \hitB{Other toe(s) amputation status (V4972)} & Other postoperative infection (99859) \\
\bottomrule
\end{tabularx}
\end{minipage}
\end{adjustbox}
\label{tab:heat_hits2}
\end{table}

After model training, we evaluated CLDD using standard ranking metrics, including precision, recall, NDCG, hit rate, and AUC, across $61{,}191$ valid patients by comparing the model's predictions with their ground-truth diseases. We then selected representative patients, as listed in Table~\ref{tab:heat_hits} and Table~\ref{tab:heat_hits2}, to quantitatively examine how CLDD prioritizes relevant diseases within its top predictions.

Among the five patients with the highest precision (Table~\ref{tab:heat_hits}), CLDD demonstrates a clear ability to rank clinically relevant diseases at the top of its predictions. For instance, patient \#15, diagnosed primarily with chronic heart failure, shows that CLDD correctly identifies multiple heart-related conditions (\textit{Chronic diastolic heart failure, Acute on chronic systolic heart failure}) within its top-5 predictions, while also retrieving comorbid status indicators such as \textit{Do not resuscitate status}. This suggests that CLDD captures latent clinical dependencies between functional and procedural diagnoses rather than relying on frequency-based correlations. Similarly, patient \#15204 presents a liver-related disease pattern, where CLDD precisely recovers both \textit{Cirrhosis of liver} and \textit{Esophageal varices}—two pathophysiologically linked disorders arising from portal hypertension—within its top-2 predictions, highlighting the model's capacity to recognize organ-specific comorbidity structures.

Beyond precision-oriented cases, we further examined patients with the highest recall scores (Table~\ref{tab:heat_hits2}), focusing on those with at least five diseases in their test sets to ensure stable evaluation. The results reveal CLDD’s capability to comprehensively recover multiple true diseases from limited historical evidence. For instance, patient \#9260 exhibits a complex pattern of diabetic complications, where the model successfully recalls several clinically consistent conditions, including \textit{osteomyelitis}, \textit{bone involvement in metabolic diseases}, and \textit{toe amputation status}. Similarly, patient \#30629 presents cardiovascular comorbidities, and CLDD accurately predicts \textit{subendocardial infarction} and \textit{chronic ischemic heart disease}, both reflecting downstream manifestations of coronary pathology.

Taken together, this case study illustrates the strong interpretability and clinical reliability of CLDD. By accurately prioritizing latent diseases from patient-disease relational patterns, CLDD enables clinicians to narrow diagnostic focus and reduce unnecessary testing. This capability not only validates the model's predictive precision and coverage at the individual level but also underscores its potential for integration into real-world, test-efficient clinical decision-support systems.



\subsection{Interpretable Patient-Disease Manifolds Learned by CLDD}
\begin{figure}[h]
    \centering
    \includegraphics[width=\linewidth]{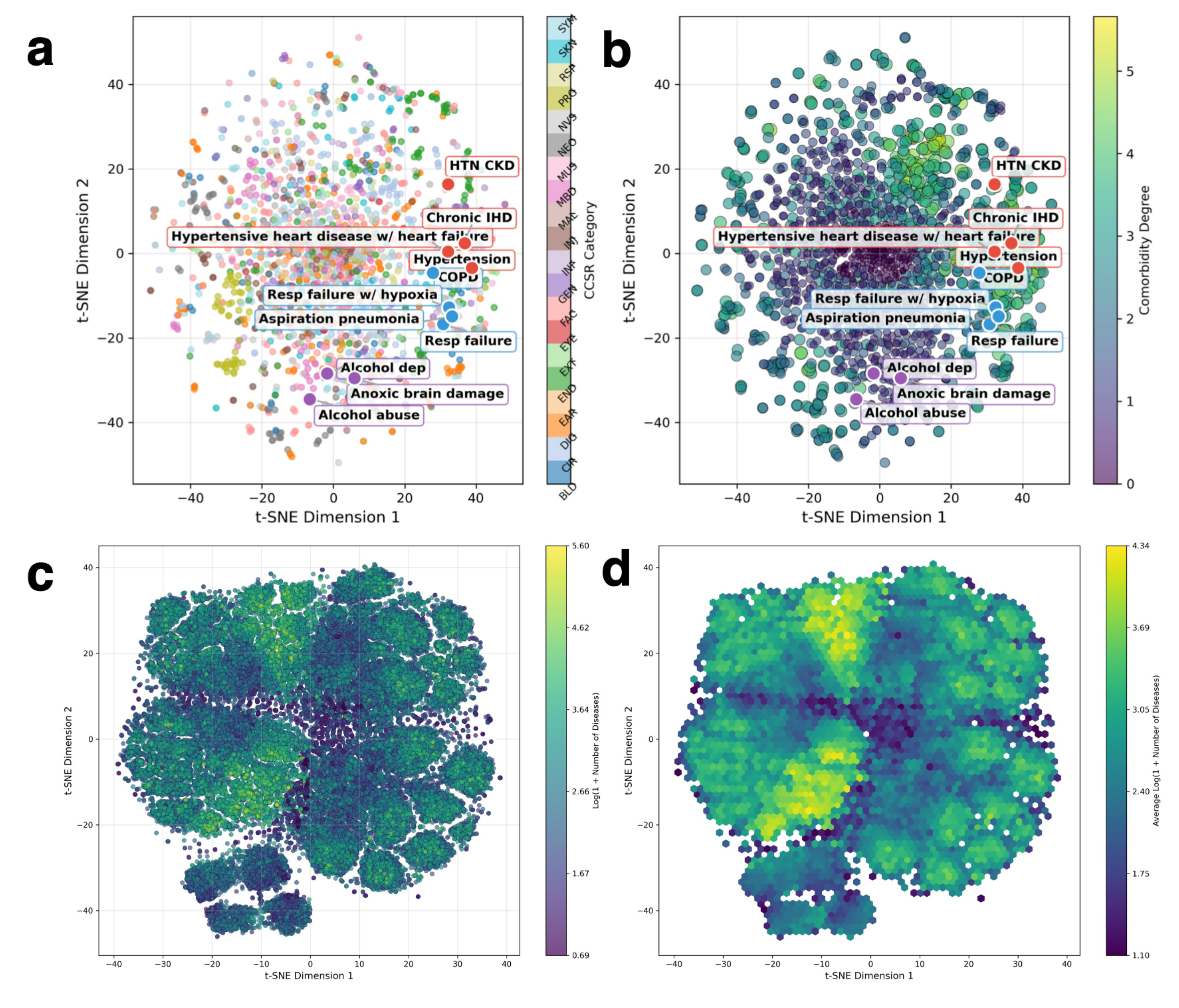}
    \caption{\textbf{Visualizations of the disease and patient embedding manifolds learned by CLDD. a,} t-SNE of disease embeddings colored by CCSR ICD-10-CM Body System. Diseases with shared physiological characteristics form coherent clusters, and representative diseases (e.g., hypertension, chronic IHD; COPD, aspiration pneumonia) form clinically interpretable local neighborhoods. \textbf{b,} t-SNE of disease embeddings colored by multimorbidity levels (log scale). High-comorbidity diseases occupy periphery regions of the manifold, whereas rare or weakly connected diseases lie near the center. \textbf{c,} t-SNE of 61,191 patient embeddings colored by disease count (log scale). Patient manifold displays clear phenotype clusters, with high disease count patients (yellow) forming dense substructures, indicating that CLDD learns meaningful patient-level representations. \textbf{d,} Density map of the patient embedding manifold, colored by average disease count per region.} 
    \label{embedding}
\end{figure}

\noindent To further examine whether CLDD learns clinically meaningful structures and captures high-order connectivity across patients and diseases, we visualize the final learned disease and patient embeddings (i.e., $z_d$ and $z_p$ in \eqref{eq11}) using t-SNE \cite{maaten2008visualizing} (Fig. \ref{embedding}). 

MIMIC-IV includes both ICD-9 and ICD-10 diagnosis codes. We first convert all ICD-9 codes into ICD-10 using General Equivalence Mappings (GEMs) \cite{centers2017icd} developed by Centers for Medicare \& Medicaid Services (CMS). We then map all ICD-10 codes to 23 Clinical Classifications Software Refined (CCSR) body systems to assign colors to t-SNE scatter plot. A detailed description of CCSR body systems is provided in Supplementary Information (Supplementary Table 4).

\textbf{Disease Embedding Space.} Fig. \ref{embedding}a shows that diseases with similar CCSR body systems tend to organize into clinically coherent clusters. We highlighted several representative diseases from distinct CCSR body systems. The cardiometabolic group (e.g., \textit{I10 Essential (primary) hypertension}, \textit{I110 Hypertensive heart disease with heart failure}, \textit{I120 Hypertensive chronic kidney disease with stage 5 chronic kidney disease or end stage renal disease}, \textit{I2510 Atherosclerotic heart disease of native coronary artery without angina pectoris}) forms tightly within Diseases of the circulatory system (CIR). These diseases share well-established physiological pathways and frequently co-occur in clinical practice. CLDD correctly embeds them in close proximity, indicating that the model truly learns clinically valid comorbidity structures.

Similarly, the respiratory diseases (i.e., \textit{J449 Chronic obstructive pulmonary disease, unspecified}, \textit{J9600 Acute respiratory failure, unspecified whether with hypoxia or hypercapnia}, \textit{J9601 Acute respiratory failure with hypoxia}, \textit{J690 Pneumonitis due to inhalation of oils and essences}) appear as a coherent subcluster and fall under Diseases of the respiratory system (RSP). Notably, CLDD also uncovers latent cross-system associations. \textit{G931 Anoxic brain damage, not elsewhere classified} from Diseases of the nervous system (NVS) lies adjacent to \textit{F1010 Alcohol abuse, in remission} and \textit{F1020 Alcohol dependence, uncomplicated}) from Mental, behavioral and neurodevelopmental disorders (MBD). The three diagnoses exhibit strong real-world clinical linkage because alcohol use disorder is a major risk factor for neurologic injury and acute encephalopathy \cite{sechi2007wernicke}. CLDD successfully embeds these seemingly disparate diseases near one another, revealing latent higher-order relationships that are not directly observable from raw co-diagnosis frequencies.

High-comorbidity diseases appear on the periphery of the manifold, whereas rare or weakly connected diseases lie near the center (Fig. \ref{embedding}b). This distribution aligns with the behavior of Laplacian-based propagation in CLDD, where highly connected diseases are more strongly separated by the multi-hop evidence they propagate, while rare diseases gravitate towards regions of maximal structural uncertainty. Such geometry discovery helps with test-free disease inference, because embedding neighborhoods encode transferable relational evidence even for diseases with limited direct observation.

\textbf{Patient Embedding Space.} Fig.~\ref{embedding}c,d visualizes the embedding space of 61,191 patients. Patients form distinct phenotypic clusters without using any disease category information. Patients with a high disease burden (log-scaled number of diagnoses) consistently appear in dense regions of the manifold, while patients with only a few diagnoses form peripheral subgroups. This smooth organization suggests that CLDD learns a continuous representation of patient health states, enabling the model to infer potential diagnoses based solely on relational similarity.

\textbf{Interpretability and Impact on Test-free Disease Detection.} Taken together, these four manifold visualizations demonstrate that CLDD constructs coherent, clinically aligned embedding spaces for both diseases and patients. The strong correspondence between disease clusters and patient phenotypes indicates that CLDD captures high-order relational structures that generalize beyond observed comorbidities. The close alignment between disease clusters and patient clusters provides an explanation for CLDD's ability to detect unseen or untested diseases, even when laboratory tests or imaging data are unavailable. Consequently, these interpretable manifolds help validate CLDD can reduce reliance on costly diagnostic testing by leveraging the rich, structurally organized information already present in routine EHR records.

\begin{figure}[t]
\centering
\subfigure[]{\includegraphics[width = 0.49\linewidth, trim={0 0 0 0},clip]{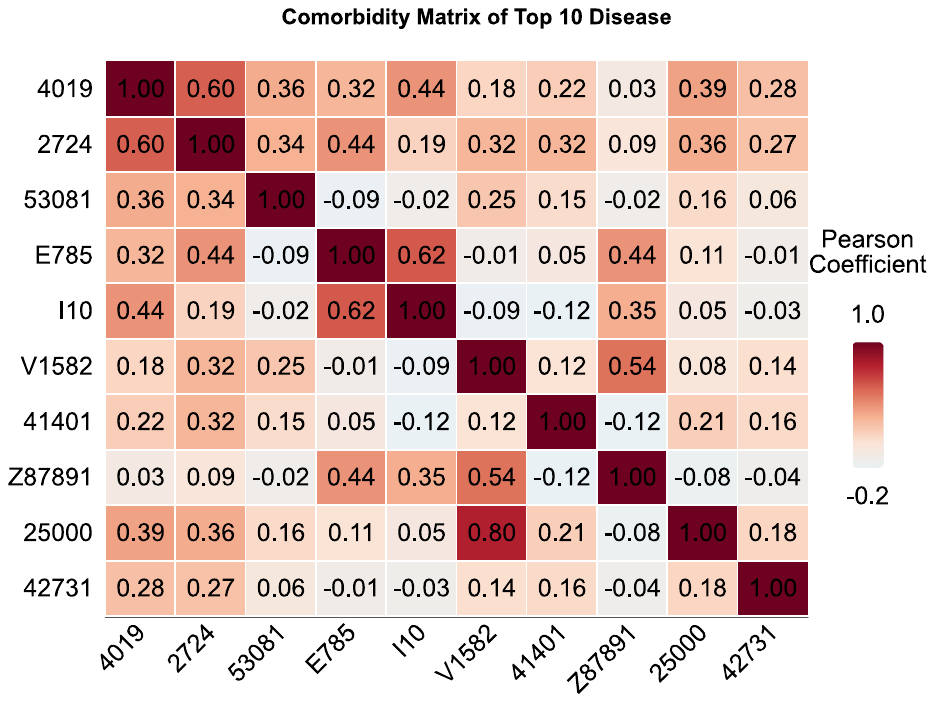}}
\hfill
\subfigure[]{\includegraphics[width = 0.49\linewidth, trim={0 0 0 0},clip]{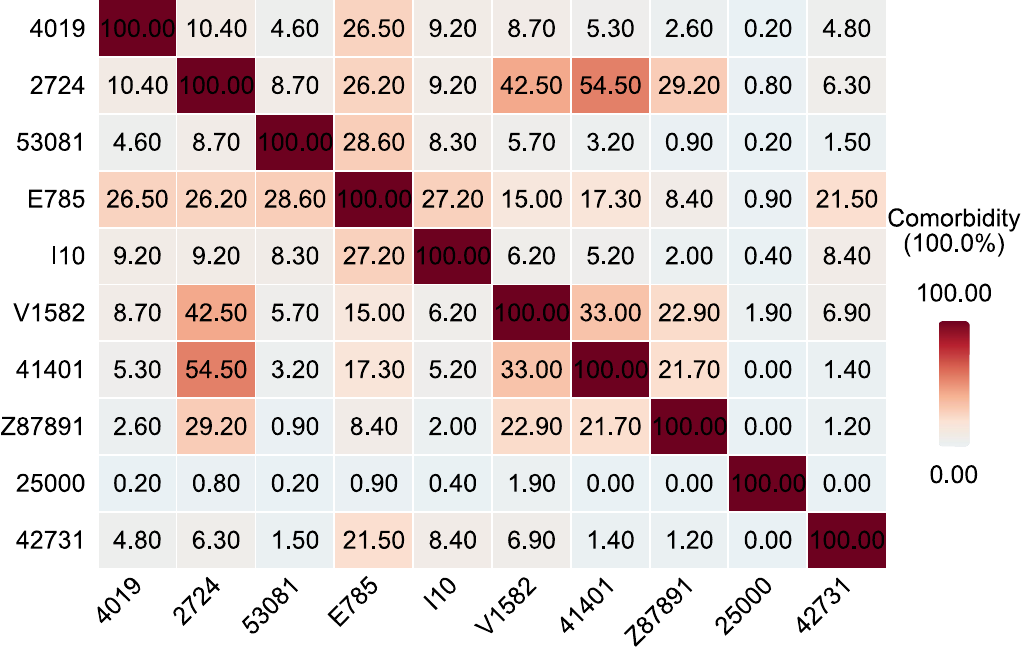}}
\subfigure[]{\includegraphics[width = 0.49\linewidth, trim={0 0 0 0},clip]{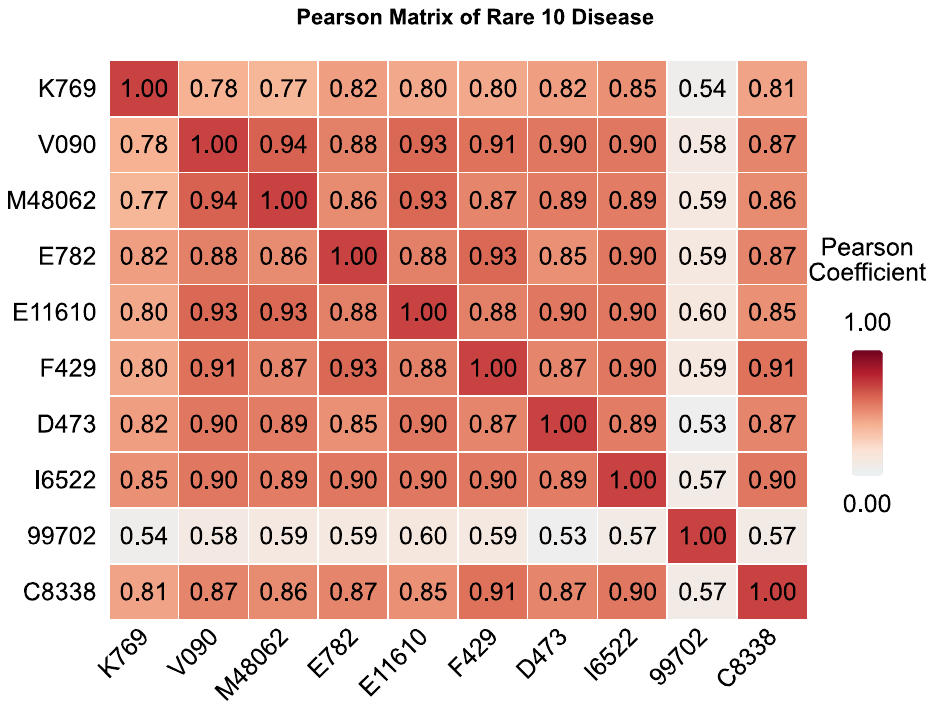}}
\hfill
\subfigure[]{\includegraphics[width = 0.49\linewidth, trim={0 0 0 0},clip]{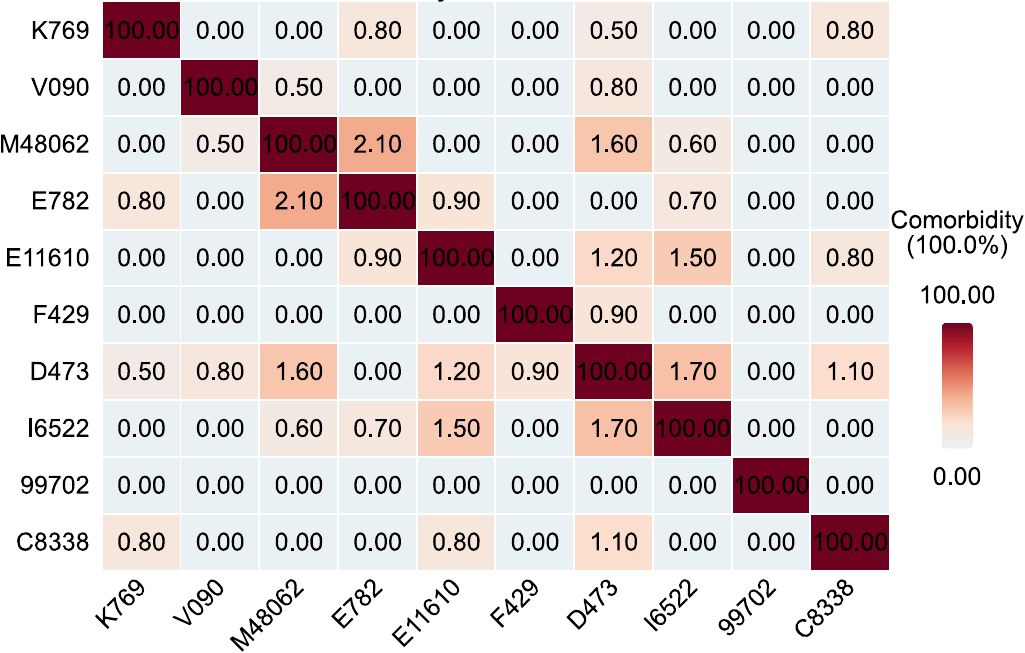}}
\caption{\textbf{Heatmap of top 10 and rare 10 discrepancy between Pearson correlation coefficient and comorbidity rate.} \textbf{a} Pearson correlation heatmap for the top 10 most frequent diseases (range from -0.2 to 1.0). \textbf{b} Comorbidity heatmap for the top 10 most frequent diseases (range from 0.00\% to 100.00\%). \textbf{c} Pearson correlation heatmap for the top 10 rarest diseases (range from 0.00 to 1.00). \textbf{d} Comorbidity heatmap for the top 10 rarest diseases (range from 0.00\% to 100.00\%).}\label{Pearson and comorbidity}
\end{figure}

\subsection{Discrepancy between Comorbidity and Pearson Correlation Coefficient of Diseases}
\noindent Our algorithmic investigation uncovers distinct patterns in the relationship between disease comorbidity rates and their Pearson correlation coefficients derived from the disease embeddings (i.e., $\boldsymbol{z}_d$ in \eqref{eq11}) learned by CLDD, reflecting heterogeneous association behaviors across disease categories. These heterogeneous association behaviors are visually quantified in Fig.~\ref{Pearson and comorbidity}, which presents four heatmaps illustrating the discrepancy between the Pearson correlation coefficient and the raw comorbidity rate for the top 10 most frequent and 10 rarest disease pairs. 


For common, acute, and readily treatable diseases, which are typically associated with higher prevalence, the discrepancy between comorbidity rates and Pearson correlation coefficients remains small. Both metrics exhibit consistent trends, indicating that statistical correlation aligns well with empirical co-diagnosis frequency in these cases.

In contrast, rare, chronic, and treatment-resistant diseases show substantial divergence between the two metrics. Such discrepancies suggest that CLDD captures latent relational structures that are not reflected in simple correlation measures, thereby uncovering hidden or indirect associations among low-prevalence conditions. These findings highlight the model’s ability to identify clinically meaningful relationships beyond traditional co-occurrence statistics, offering deeper insight into complex disease interactions and potential comorbidity mechanisms.

To quantify these discrepancies, we define a discrepancy measure as the absolute difference between comorbidity rates and Pearson correlation coefficients: 
$$\mathrm{Discrepancy} = |\mathrm{Comorbidity \ Rate} - \mathrm{Pearson \ Correlation \ Coefficient}|.$$ 
By systematically ranking all disease pairs according to this discrepancy measure, we identify those with the largest discrepancies as priority candidates for further clinical or epidemiological analysis. Representative examples of such disease pairs are provided in the Supplementary Information (Supplementary Table 5), highlighting cases where CLDD uncovers latent associations that are not apparent from empirical co-diagnosis frequencies.

To illustrate how such a discrepancy measure manifests in practice, we introduce two representative disease pairs with large discrepancy values.

\textbf{A high-comorbidity but low-correlation pair: Coronary atherosclerosis of unspecified type of vessel, native or graft (41400) and Other and unspecified hyperlipidemia (2724).} This pair exhibits a high empirical comorbidity rate of 42.97\%, yet their embedding-based Pearson correlation is only 0.0845. Clinically, hyperlipidemia is a well-known risk factor for coronary atherosclerosis and thus frequently appears together in routine cardiovascular evaluations \cite{libby2002inflammation,stone20142013}. However, the two conditions can follow diverse clinical pathways and progressions across patients, and may be co-coded due to screening protocols or chronic disease management workflows rather than because they form a tightly coupled latent phenotype. CLDD correctly assigns low embedding similarity, indicating that despite frequent co-diagnosis, the two diseases do not share strong high-order latent relations in the patient-disease graph. This illustrates how the model differentiates true latent affinity from co-occurrence driven by general preventive care or diagnostic bundling.

\textbf{A low-comorbidity but high-correlation pair: Obstructive chronic bronchitis with (acute) exacerbation (49121) and Malignant neoplasm of other parts of bronchus or lung (1628).} In contrast, this pair shows a low comorbidity rate of 6.2\%, yet a relatively high embedding-based Pearson correlation of 0.35. Although the comorbidity rate is low, the relationship between the two diseases is clinically plausible: long-term airway inflammation, smoking exposure, and shared pulmonary risk factors create overlapping patient subpopulations. These indirect links are captured by CLDD’s multi-hop propagation, as the two diseases tend to co-occur with similar intermediate pulmonary conditions (e.g., chronic airway obstruction, respiratory infections), bringing their embeddings closer even when direct co-diagnosis counts are sparse. This highlights CLDD's ability to uncover latent comorbidity structures that is underrepresented in raw EHR frequencies due to low disease prevalence or under-reporting.

Taken together, these two contrasting examples show that large values of the discrepancy measure systematically correspond to clinically interpretable mismatches between comorbidity rates and deeper relational patterns captured by CLDD, thereby supporting the validity of the discrepancy as a principled signal for prioritizing disease pairs for further analysis.


\subsection{Quantitative Comparison with Baselines}\label{metrics}
\subsubsection{Evaluation Metrics and Baselines}
\noindent For each patient in the test set, all diseases that the patient has not interacted with are treated as negative samples. Each method outputs the patient’s preference scores over all diseases, excluding the positive ones in the training set. To evaluate both the effectiveness of top-K predictions and overall ranking quality, we adopt five standard evaluation metrics: recall@K, precision@K, hit rate@K, ndcg@K, and AUC, with K = 20 by default. Reported values represent the average across all patients in the test set.

To benchmark CLDD, we compare it against five representative collaborative filtering and graph-based baselines:
\begin{itemize}
\item \textbf{MFBPR} \cite{rendle2012bpr}: This is a matrix factorization method optimized by the Bayesian personalized ranking (BPR) loss.
\item \textbf{NFM} \cite{he2017neuralfactorizationmachinessparse}: This method extends matrix factorization by replacing dot products with neural layers to capture nonlinear feature interactions.
\item \textbf{NGCF} \cite{wang2019neural}: A graph convolutional model that propagates embeddings through higher-order connectivity in the user–item (patient–disease) graph. 
\item \textbf{LightGCN} \cite{he2020lightgcnsimplifyingpoweringgraph}: A lightweight version of NGCF that removes transformation and nonlinearity layers while retaining strong representational power.
\item \textbf{KGAT} \cite{wang2019kgat}: Incorporates attention mechanisms and augments graph learning with external knowledge-graph information.
\end{itemize}

\subsubsection{Performance Analysis}\label{subsec5}
\noindent Table~\ref{tab1} summarizes the performance comparisons of CLDD and five representative baselines on the MIMIC-IV dataset. Across all five evaluation metrics, CLDD consistently achieves superior performances, demonstrating its strong capability in disease detection and patient-disease relationship modeling.

Specifically, CLDD attains an AUC of 0.7235, outperforming the strongest baseline by \textbf{2.94\%}. Substantial improvements are also observed on Recall (+6.33\%), Precision (+7.63\%), NDCG (+6.87\%), and Hit Rate (+7.64\%). These consistent gains indicate that CLDD not only enhances the overall ranking quality but also effectively identifies true disease associations at top positions, enabling more efficient test prioritization for patients. 

Compared with traditional matrix factorization methods such as MFBPR and NFM, CLDD yields significant performance gains, underscoring the importance of relational reasoning over purely latent-factor modeling. Among graph learning baselines, NGCF and LightGCN already leverage structural information and perform competitively, yet CLDD further advances performance by introducing collaborative propagation with adaptive hop-wise weighting and patient-specific priors. This design mitigates over-smoothing and noise accumulation, leading to more discriminative and robust embeddings. Ablations on the feature matrix (Supplementary Table 1) and the number of embedding propagation layers (Supplementary Table 2) are provided in the Supplementary Information file.

Overall, the consistent superiority across multiple metrics confirms the effectiveness of CLDD’s collaborative graph learning framework. By capturing both direct and higher-order dependencies among patients and diseases, CLDD establishes a strong foundation for test-free disease detection in large-scale EHR data.

\begin{table}[h]
\caption{\textbf{Performance comparison of CLDD and baseline models on the MIMIC-IV dataset.} CLDD consistently outperforms all baselines across all metrics. The last row reports relative percentage improvements of CLDD over the strongest baseline. Experiments for CLDD are repeated ten times and the mean is reported.}\label{tab1}%
\begin{tabular}{@{}llllll@{}}
\toprule
Models    & AUC & Recall & Precision & NDCG & Hit Rate\\
\midrule
MFBPR      &  0.5010  & 0.0670  & 0.0099 & 0.0324 & 0.1888 \\
NFM      &  0.5947  &  0.0384 & 0.0067 & 0.0384 & 0.1222\\
LightGCN   & \underline{0.7029}   & 0.0568  & 0.0081 & 0.0261 & 0.1374 \\
NGCF       & 0.6668   & \underline{0.0743}  & \underline{0.0118} & \underline{0.0393} & \underline{0.1977} \\
KGAT       &  0.6559  & 0.0603  & 0.0092 & 0.0295 & 0.1580 \\
\textbf{CLDD(Ours)} & \textbf{0.7235}   & \textbf{0.079}  & \textbf{0.0127} & \textbf{0.042} & \textbf{0.2128} \\

\midrule
\%Improv. & 2.94\% & 6.33\%  &	7.63\%	& 6.87\%  &	7.64\% \\
\botrule
\end{tabular}
\end{table}

\section{Discussion}\label{sec3}
\noindent We propose \textbf{CLDD}, a collaborative graph-based framework that detects potential diseases without relying on exhaustive medical tests by leveraging large-scale EHR data. We formulate disease detection as a patient-disease collaborative learning problem, grounded in the observation that patients with similar clinical profiles often share disease patterns. By introducing a novel embedding propagation mechanism, CLDD integrates local relational evidence and higher-order comorbidity structures, thereby enhancing the robustness and discriminative power of learned representations.

Our empirical evaluation on the MIMIC-IV dataset establishes CLDD as a strong baseline for EHR-based disease detection. To the best of our knowledge, this work is the first to systematically benchmark collaborative filtering approaches on MIMIC-IV. CLDD achieves consistently superior results across five metrics compared to previous methods. For high-precision patients, CLDD successfully recovers more than half of the masked diseases within a short prediction list (Top-20). In high-recall cases, the model attains perfect recovery of three masked conditions (recall rate 1.0). These findings indicate that CLDD effectively identifies latent disease risks, offering potential to reduce unnecessary diagnostic tests and improve early-stage screening.

Beyond performance, CLDD contributes conceptually by bridging recommender systems and clinical informatics. The propagation-based design enables the model to infer disease likelihood from collective patient-disease interactions, complementing traditional test-driven diagnostics. In practical terms, this collaborative learning perspective can assist physicians by providing probabilistic disease candidates for further verification, thereby accelerating differential diagnosis and supporting resource-efficient medical decision-making.

Nevertheless, several limitations require further exploration. First, our experiments rely on a single-center dataset (MIMIC-IV), which may limit generalizability to other healthcare systems with different demographic or coding distributions. Second, while CLDD captures relational structure effectively, its interpretability at the feature or patient level remains limited. Future work may incorporate attention or causal graph modules to trace the contribution of specific patient attributes or clinical pathways. Third, CLDD currently models only structured EHR entries (diagnoses, prescriptions, procedures). Extending the framework to multimodal data—including clinical notes, laboratory tests, and imaging—could further enhance predictive coverage and clinical relevance.

In conclusion, CLDD provides a scalable and interpretable foundation for collaborative disease detection. By integrating patient–disease correlations through deep graph propagation, our approach opens new directions for test-free, data-driven healthcare systems and contributes to the broader goal of intelligent, cost-efficient medical diagnosis.



\section{Methods}\label{sec4}

\subsection{Data and Processing}\label{process}
\noindent We conducted experiments on the publicly available MIMIC-IV v2.2 \cite{mimiciv_v2} dataset, which contains de-identified electronic health records (EHR) from the Beth Israel Deaconess Medical Center. Specifically, we utilized the Hosp module, which comprises comprehensive hospital-wide EHR data including diagnoses, prescriptions, and procedures. Our data preprocessing follows the methodologies employed in ManyDG \cite{yang2023manydg}, GAMENet \cite{shang2019gamenet} and SafeDrug \cite{yang2021safedrug}. A summary of the processed dataset statistics is provided in Supplementary Information file (Supplementary Table 3). A visualization of the statistics is shown in Supplementary Fig. 1.


\subsubsection*{Patient selection}
\noindent To ensure the reliability and completeness of patient records, we applied two filtering criteria. First, only patients with at least two hospital admissions were retained to provide sufficient longitudinal information for model training. Second, we included only those who had records in all three modalities---\textit{prescriptions}, \textit{diagnoses}, and \textit{procedures}---so that each patient's medical profile was both complete and consistent. These records were extracted from "prescriptions.csv", "diagnoses\_icd.csv", and "procedures\_icd.csv" files. Moreover, medications from "prescriptions.csv" were mapped from NDC node to RxCUI identifiers and subsequently to ATC classifications, truncated to level~3 (ATC3) to balance granularity and robustness. We then restricted to the 300 most prevalent ATC3 categories to focus on well-represented pharmacotherapies. To enhance data completeness and facilitate potential cross-database alignment, DrugBank annotations were integrated to retrieve up to three canonical SMILES for each ATC3 category, and classes lacking structural coverage were excluded.

\subsubsection*{Disease selection}
\begin{figure}
\centering
\subfigure[]{\includegraphics[width = 0.49\linewidth, trim={0 -15 0 0},clip]{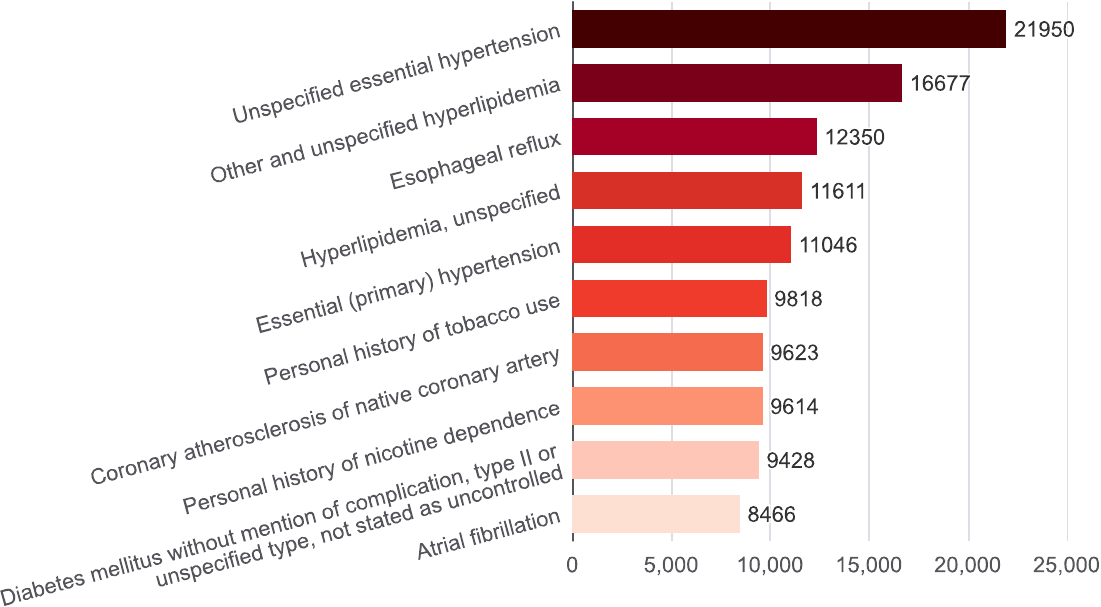}}
\hfill
\subfigure[]{\includegraphics[width = 0.49\linewidth, trim={0 0 0 0},clip]{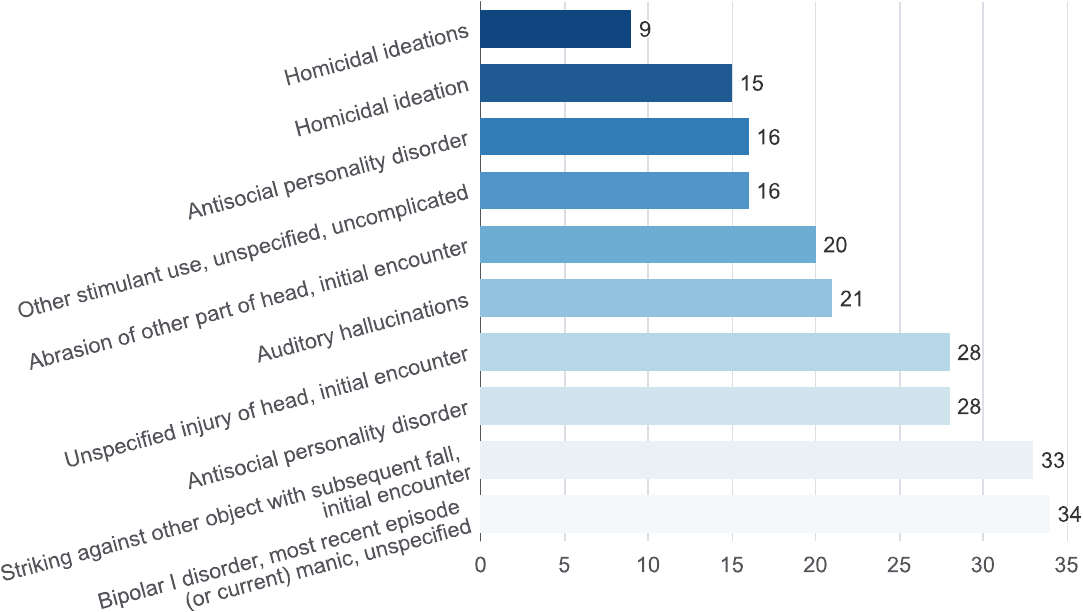}}
\caption{\textbf{Distribution of disease prevalence frequencies.} \textbf{a} Frequency counts and corresponding names (using the full English nomenclature from the ICD coding system) of the top 10 most prevalent diseases in the dataset. \textbf{b} Frequency counts and corresponding names (using the full English nomenclature from the ICD coding system) for the 10 least prevalent diseases.} \label{top 10 and bottom 10}
\end{figure}

\noindent Each disease in MIMIC-IV is identified by a unique ICD code. To mitigate data sparsity, we count the frequency of each disease across all patients and retain the top 2,000 most frequent diseases. Among these selected diseases, the 10 most and least common diseases are listed in Fig. \ref{top 10 and bottom 10}. The resulting filtered dataset provides a more representative patient-disease interaction graph.

\subsubsection*{Dataset construction and splitting}
\noindent Based on the filtered data, we construct a bipartite adjacency matrix between patients and diseases, as well as a feature matrix capturing demographic attributes such as age, gender, and race. The feature encoding details follow Section~\ref{Embedding}.
For temporal splitting, each patient's disease interaction sequence is ordered chronologically. The first 80\% of interactions are used for training and the remaining 20\% for testing, with no separate validation set. This 8:2 ratio ensures that the model is evaluated on future, unseen interactions, aligning with the predictive nature of disease detection.

\subsubsection*{Feature encoding}
\noindent The extracted demographic features include age, gender, and race, each encoded via one-hot representations. Due to the de-identifying protocol in MIMIC-IV, the anchor age represents the patient's age in a shifted reference year rather than their true age. When the anchor age exceeds 89, it is replaced with 91 to preserve anonymity. We divide the age interval into nine mutually exclusive groups: [18, 20], [21, 30], [31, 40], ..., [81, 89], and [91]. Gender is encoded as 0 for male and 1 for female. Race includes 33 categories, each represented by a unique one-hot vector. These demographic vectors are concatenated with the learnable embedding components introduced in Section~\ref{Embedding}.

\subsection{Framework of the Model}\label{framework}
\noindent We propose CLDD, a graph-based collaborative learning framework for disease detection. The model comprises four components: (1) an embedding layer that initializes the patient and disease embeddings; (2) a neighbor aggregation mechanism that encodes one-hop relational evidence; (3) a stack of embedding propagation layers that aggregate higher-order information of the central node; (4) a prediction layer that calculate the predicted score of patient-disease pairs.

\subsubsection{Embedding Layer}
\label{Embedding}
\noindent Each disease $d$ is represented by a learnable embedding vector $\boldsymbol{z}_d\in\mathbb{R}^{k}$. Each patient $p$ is represented as a concatenation of a learnable component and a fixed attribute component:
$$\boldsymbol{z}_p=[\boldsymbol{z}_{p^{+}};\boldsymbol{z}_{p^{-}}]\in\mathbb{R}^{k},\ \ \boldsymbol{z}_{p^{+}}\in\mathbb{R}^{k-f}, \ \ \boldsymbol{z}_{p^{-}}\in\mathbb{R}^{f},$$ 
where $\boldsymbol{z}_{p^{+}}$ is learnable, and $\boldsymbol{z}_{p^{-}}$ encodes fixed demographic attributes such as gender, age, and race, with $0\leq f<k$. 
Stacking all disease and patient embeddings yield the initial embedding matrix $\boldsymbol{Z}^{(0)} \in \mathbb{R}^{(P+D)\times k}$.

These representations are subsequently refined through messages propagation over the patient-disease interaction graph $\mathcal{G}=(\mathcal{P}\cup\mathcal{D},\mathcal{E})$. This graph-based refinement explicitly injects collaborative signals into the embeddings, generating more discriminative and noise-resilient representations that improve downstream disease-detection performance.

\subsubsection{Neighbor Aggregation}
\noindent The patients linked to a disease can be interpreted as its features and used to measure the collaborative similarity of two diseases. The construction is symmetric for patients. For clarity, we describe aggregation from diseases to patients. 

Let $\mathcal{N}_p$ and $\mathcal{N}_d$ denote the one-hop neighbors of patient $p$ and disease $d$, respectively. For an edge $(p, d) \in \mathcal{E}$, the message from disease $d$ to patient $p$ is defined as:
\begin{equation}\label{eq1}
\boldsymbol{I}_{p\leftarrow d}=f(\boldsymbol{z}_p,\boldsymbol{z}_d, w_{pd}),
\end{equation}
where $\boldsymbol{I}_{p\leftarrow d}$ is the message embedding, denoting the information to be propagated. $f(\cdot)$ is the information encoding function, which takes $z_p$ and $z_d$ as input, and use a coefficient $w_{pd}$ to control the decay factor on each propagation on edge $(p,d)$.
We define $f(\cdot)$ as a weighted sum aggregator parameterized by a learnable matrix $\boldsymbol{W}$:
\begin{equation}
    \boldsymbol{I}_{p\leftarrow d}=\frac{1}{\sqrt{|\mathcal{N}_p||\mathcal{N}_d|}} \boldsymbol{W} (\boldsymbol{z}_d \circ \boldsymbol{z}_p),
\end{equation}
where $\boldsymbol{W} \in \mathbb{R}^{k'\times k}$ is a learnable transformation, and $k'$ is the transformation size. Term $\boldsymbol{z}_d \circ \boldsymbol{z}_p$ denote the interaction between $\boldsymbol{z}_d$ and $\boldsymbol{z}_p$, where $\circ$ denote the element-wise multiplication. Following the work of graph convolution network \cite{kipf2017semisupervised}, we set $w_{pd}$=$1/\sqrt{|\mathcal{N}_p||\mathcal{N}_d|}$, which can be viewed as the discount factor, considering the weight decay of the message being passed along the path. 

The aggregated neighborhood information is then used to update each patient's embedding:
\begin{equation}\label{eq3}
\boldsymbol{z}_p^{(1)} = \text{LeakyReLU}\left(\sum_{d \in \mathcal{N}_p} \boldsymbol{I}_{p\leftarrow d}\right),
\end{equation}
where $\boldsymbol{z}_p^{(1)}$ is the refined embedding of patient $p$ obtained after the first embedding propagation layer. The LeakyReLU \cite{maas2013rectifier} activation function allows information to encode both positive and small negative signals, mitigating inactive neuron effects. This operation effectively aggregates first-order neighborhood signals, allowing each patient to integrate diagnostic evidence from related diseases. Analogously, $\boldsymbol{z}_d^{(1)}$ is computed by aggregating information from all connected patients. Overall, neighbor aggregation explicitly captures first-order connectivity information and establishes the foundation for higher-order propagation.

\subsubsection{Embedding Propagation Layer}
\noindent Real-world patient-disease graphs often contain informative multi-hop structure. Patients connected through shared diseases reveal comorbidity patterns, while overlapping diseases expose common etiologies. To exploit such structure, CLDD refines node representations through multi-hop propagation and adaptive hop-mixing.

We fist define the patient-disease adjacency matrix $\boldsymbol{A}$:
\begin{equation}\label{eq:A}
\boldsymbol{A} \;=\;
\begin{bmatrix}
\boldsymbol{0} & \boldsymbol{Y} \\
\boldsymbol{Y}^{\!\top} & \boldsymbol{0}
\end{bmatrix}
\in \mathbb{R}^{(P+D)\times(P+D)}.
\end{equation}

The symmetrically normalized Laplacian matrix $\boldsymbol{\hat{A}}$ is given by:
\begin{equation}\label{eq7}
\boldsymbol{\hat{A}}=\boldsymbol{D}^{-\frac{1}{2}} \boldsymbol{\widetilde{A}} \boldsymbol{D}^{-\frac{1}{2}} \ \rm{and} \ \boldsymbol{\widetilde{A}} = \boldsymbol{A} + \boldsymbol{I} =
\left[\begin{array}{cc}
\boldsymbol{I}   &  \boldsymbol{Y}  \\
\boldsymbol{Y}^T &  \boldsymbol{I}
\end{array}\right].
\end{equation}

With the information aggregated from the first-hop neighbors, the representation of the node is further updated through iterative propagation. We can stack embedding propagation layers to explore the high-order connectivity information.

Empirically, single-hop propagation may fail to capture long-range dependencies, while excessively deep diffusion risks oversmoothing. To balance locality and globality, CLDD linearly combines signals from multiple hop distances, adaptively weighting each via a layer-wise softmax. Given the layer input \(\boldsymbol{Z}^{(l-1)}\in\mathbb{R}^{(P+D)\times d_{l-1}}\), the hop-mixed neighborhood signal $\boldsymbol{S}^{(l)}$ is formulated as:
\begin{equation}\label{eq:Sl}
\boldsymbol{S}^{(l)}
= \Big(\sum_{i=1}^K \beta_i^{(l)} \boldsymbol{\hat{A}}^{i}\Big)\boldsymbol{Z}^{(l-1)},
\end{equation}
with nonnegative mixture weights obtained via a per-layer softmax:
\begin{equation}\label{eq:beta}
\beta^{(l)}_{i}=
\frac{\exp(\alpha^{(l)}_{i})}{\sum_{j=1}^K\exp(\alpha^{(l)}_{j})},\quad
i=1,...,K,\quad
\sum_{i=1}^K\beta^{(l)}_{i}=1,
\end{equation}
where $K$ denotes the maximum hop order. This construction lets the network emphasise short-range evidence when local cohorts are diagnostic, while leveraging longer paths when disease signals are diffuse.

To jointly capture additive neighbourhood evidence and multiplicative state-context interactions, each layer fuses two complementary branches before activation. The \emph{linear} branch performs a graph convolution:
\begin{equation}\label{eq:Gl}
\boldsymbol{G}^{(l)}=\boldsymbol{S}^{(l)}\boldsymbol{W}^{(l)}_{\mathrm{gc}}+\boldsymbol{1}\big(\boldsymbol{b}^{(l)}_{\mathrm{gc}}\big)^{\!\top},
\end{equation}
while the \emph{bilinear} branch modulates the aggregated signal by the current node state via a Hadamard product,
\begin{equation}\label{eq:Bl}
\boldsymbol{B}^{(l)}=\big(\boldsymbol{Z}^{(l-1)}\circ\boldsymbol{S}^{(l)}\big)\boldsymbol{W}^{(l)}_{\mathrm{bi}}+\boldsymbol{1}\big(\boldsymbol{b}^{(l)}_{\mathrm{bi}}\big)^{\!\top},
\end{equation}
with \(\boldsymbol{W}^{(l)}_{\mathrm{gc}},\boldsymbol{W}^{(l)}_{\mathrm{bi}}\in\mathbb{R}^{d_{l-1}\times d_{l}}\) and \(\boldsymbol{b}^{(l)}_{\mathrm{gc}},\boldsymbol{b}^{(l)}_{\mathrm{bi}}\in\mathbb{R}^{d_l}\).
The sum of both branches is passed through a LeakyReLU activation, dropout with rate \(\rho^{(l)}\), and row-wise \(\ell_2\) normalization to stabilize training and control feature scale. The final representation $\boldsymbol{Z}^{(l)}$ can be represented as:
\begin{align}
\boldsymbol{Z}^{(l)}&=\mathrm{RowNorm}_2\big(\mathrm{Dropout}\!\big(\mathrm{LeakyReLU}\big(\boldsymbol{G}^{(l)}+\boldsymbol{B}^{(l)}\big);\rho^{(l)}\big)\big),\label{eq:Hl}
\end{align}
Equation \eqref{eq:Sl}-\eqref{eq:Hl} together define the layer-wise update \(\boldsymbol{Z}^{(l)}=\mathrm{Func}\!\big(\boldsymbol{Z}^{(l-1)}\big)\), enabling integration of information from multiple hop distances without loss of local specificity. This propagation can simultaneously update the representations for all patients and diseases in $O(|\mathcal{E}|\,d_ld_{l-1})$ time per hop, ensuring efficient and scalable end-to-end training.

\subsubsection{Prediction Layer}
\noindent After going through the embedding propagation layer, we obtain embeddings in $L$ layers. Embeddings from different layers capture information at distinct levels, concatenating them yields a comprehensive representation. We concatenate the embeddings in $L$ layers to get the final embedding for model prediction. Aggregating them in this way can fully utilize the information. Also, concatenation is the simplest way to aggregate the embeddings. This method is commonly used in the domain of GCNs \cite{wang2019neural}. The final embedding for prediction can be represented as:

\begin{equation}\label{eq10}
\boldsymbol{Z} = \boldsymbol{Z}^{(0)}\ || \ \boldsymbol{Z}^{(1)}\ || \ ...\ ||\ \boldsymbol{Z}^{(L)}.
\end{equation}

Finally, we use the inner product to get the predicted score of patient towards a certain disease:
\begin{equation}\label{eq11}
\hat{x}_{pd} = \boldsymbol{z}_p^T \boldsymbol{z}_d.
\end{equation}

Other, more complicated choices, such as neural network-based interaction functions \cite{he2017neural}, are left for exploration in future work.

\subsection{Optimization}\label{optimization}
\noindent To optimize the model parameters, we employ the BPR loss \cite{rendle2012bpr} as objective:
\begin{equation}\label{eq12}
L_{BPR}=\sum_{(p,d_1,d_2)\in D_S} -\ln \sigma(\hat{x}_{pd_1} - \hat{x}_{pd_2}) + \lambda||\boldsymbol{\Theta}||^2,
\end{equation}
where $D_S := \{(p,d_1,d_2)|d_1 \in D^+ \wedge d_2 \in D^-\}$ denotes the observed training dataset where patient $p$ is assumed to be more likely to have disease $d_1$ than $d_2$. $D^+$ denotes the positive disease set and $D^-$ denotes the negative disease set. $\lambda$ is model regularization parameter, and $\boldsymbol{\Theta}$ denotes all trainable model parameters, i.e. $\boldsymbol{\Theta} = \{ \boldsymbol{Z}_{p^+}, \boldsymbol{Z}_{d}, \boldsymbol{\alpha}, \{\boldsymbol{W}^{(l)}_{gc},\boldsymbol{b}^{(l)}_{gc},\boldsymbol{W}^{(l)}_{bi},\boldsymbol{b}^{(l)}_{gc}\}^L_{l=1} \}$, where $\boldsymbol{\alpha}=\{\alpha_i^{(l)}: i=1,\ldots,K,l=1,\ldots,L\}$. We use mini-batch Adam to optimize the model and update model parameters. Detailed parameter settings of CLDD are provided in the Supplementary Information file. BPR loss is commonly used in recommender systems \cite{rendle2012bpr, wang2019neural}. It is designed to handle implicit feedback, where the goal is to rank observed items higher than non-observed ones. In particular, for a batch of triples $(p,d_1,d_2) \in D_S$, we put them into the embedding propagation layer and update model parameters using the gradients of the loss function.

\section*{Data availability}
\noindent The MIMIC-IV v2.2 dataset that support the findings of this study is available at https://physionet.org/content/mimiciv/2.2/ with the identifier DOI: 10.13026/kpb9-mt58. Only credentialed users who sign the DUA can access the files.


\section*{Code availability}
\noindent The proposed CLDD model and Python scripts to reproduce results reported in this work are available on Github (https://github.com/haokun-zhao/Collaborative-Disease-Detection).

\bibliography{sn-bibliography}


\begin{thebibliography}{31}
\ifx \bisbn   \undefined \def \bisbn  #1{ISBN #1}\fi
\ifx \binits  \undefined \def \binits#1{#1}\fi
\ifx \bauthor  \undefined \def \bauthor#1{#1}\fi
\ifx \batitle  \undefined \def \batitle#1{#1}\fi
\ifx \bjtitle  \undefined \def \bjtitle#1{#1}\fi
\ifx \bvolume  \undefined \def \bvolume#1{\textbf{#1}}\fi
\ifx \byear  \undefined \def \byear#1{#1}\fi
\ifx \bissue  \undefined \def \bissue#1{#1}\fi
\ifx \bfpage  \undefined \def \bfpage#1{#1}\fi
\ifx \blpage  \undefined \def \blpage #1{#1}\fi
\ifx \burl  \undefined \def \burl#1{\textsf{#1}}\fi
\ifx \doiurl  \undefined \def \doiurl#1{\url{https://doi.org/#1}}\fi
\ifx \betal  \undefined \def \betal{\textit{et al.}}\fi
\ifx \binstitute  \undefined \def \binstitute#1{#1}\fi
\ifx \binstitutionaled  \undefined \def \binstitutionaled#1{#1}\fi
\ifx \bctitle  \undefined \def \bctitle#1{#1}\fi
\ifx \beditor  \undefined \def \beditor#1{#1}\fi
\ifx \bpublisher  \undefined \def \bpublisher#1{#1}\fi
\ifx \bbtitle  \undefined \def \bbtitle#1{#1}\fi
\ifx \bedition  \undefined \def \bedition#1{#1}\fi
\ifx \bseriesno  \undefined \def \bseriesno#1{#1}\fi
\ifx \blocation  \undefined \def \blocation#1{#1}\fi
\ifx \bsertitle  \undefined \def \bsertitle#1{#1}\fi
\ifx \bsnm \undefined \def \bsnm#1{#1}\fi
\ifx \bsuffix \undefined \def \bsuffix#1{#1}\fi
\ifx \bparticle \undefined \def \bparticle#1{#1}\fi
\ifx \barticle \undefined \def \barticle#1{#1}\fi
\bibcommenthead
\ifx \bconfdate \undefined \def \bconfdate #1{#1}\fi
\ifx \botherref \undefined \def \botherref #1{#1}\fi
\ifx \url \undefined \def \url#1{\textsf{#1}}\fi
\ifx \bchapter \undefined \def \bchapter#1{#1}\fi
\ifx \bbook \undefined \def \bbook#1{#1}\fi
\ifx \bcomment \undefined \def \bcomment#1{#1}\fi
\ifx \oauthor \undefined \def \oauthor#1{#1}\fi
\ifx \citeauthoryear \undefined \def \citeauthoryear#1{#1}\fi
\ifx \endbibitem  \undefined \def \endbibitem {}\fi
\ifx \bconflocation  \undefined \def \bconflocation#1{#1}\fi
\ifx \arxivurl  \undefined \def \arxivurl#1{\textsf{#1}}\fi
\csname PreBibitemsHook\endcsname

\bibitem[\protect\citeauthoryear{Bossuyt et~al.}{2012}]{bossuyt2012beyond}
\begin{barticle}
\bauthor{\bsnm{Bossuyt}, \binits{P.M.}},
\bauthor{\bsnm{Reitsma}, \binits{J.B.}},
\bauthor{\bsnm{Linnet}, \binits{K.}},
\bauthor{\bsnm{Moons}, \binits{K.G.}}:
\batitle{Beyond diagnostic accuracy: the clinical utility of diagnostic tests}.
\bjtitle{Clinical chemistry}
\bvolume{58}(\bissue{12}),
\bfpage{1636}--\blpage{1643}
(\byear{2012})
\end{barticle}
\endbibitem

\bibitem[\protect\citeauthoryear{Qaseem et~al.}{2012}]{qaseem2012appropriate}
\begin{botherref}
\oauthor{\bsnm{Qaseem}, \binits{A.}},
\oauthor{\bsnm{Alguire}, \binits{P.}},
\oauthor{\bsnm{Dallas}, \binits{P.}},
\oauthor{\bsnm{Feinberg}, \binits{L.E.}},
\oauthor{\bsnm{Fitzgerald}, \binits{F.T.}},
\oauthor{\bsnm{Horwitch}, \binits{C.}},
\oauthor{\bsnm{Humphrey}, \binits{L.}},
\oauthor{\bsnm{LeBlond}, \binits{R.}},
\oauthor{\bsnm{Moyer}, \binits{D.}},
\oauthor{\bsnm{Wiese}, \binits{J.G.}}, et al.:
Appropriate use of screening and diagnostic tests to foster high-value,
  cost-conscious care.
American College of Physicians
(2012)
\end{botherref}
\endbibitem

\bibitem[\protect\citeauthoryear{Yang et~al.}{2021}]{yang2021safedrug}
\begin{botherref}
\oauthor{\bsnm{Yang}, \binits{C.}},
\oauthor{\bsnm{Xiao}, \binits{C.}},
\oauthor{\bsnm{Ma}, \binits{F.}},
\oauthor{\bsnm{Glass}, \binits{L.}},
\oauthor{\bsnm{Sun}, \binits{J.}}:
Safedrug: Dual molecular graph encoders for recommending effective and safe
  drug combinations.
arXiv preprint arXiv:2105.02711
(2021)
\end{botherref}
\endbibitem

\bibitem[\protect\citeauthoryear{Shang et~al.}{2019}]{shang2019gamenet}
\begin{bchapter}
\bauthor{\bsnm{Shang}, \binits{J.}},
\bauthor{\bsnm{Xiao}, \binits{C.}},
\bauthor{\bsnm{Ma}, \binits{T.}},
\bauthor{\bsnm{Li}, \binits{H.}},
\bauthor{\bsnm{Sun}, \binits{J.}}:
\bctitle{Gamenet: Graph augmented memory networks for recommending medication
  combination}.
In: \bbtitle{Proceedings of the AAAI Conference on Artificial Intelligence},
vol. \bseriesno{33},
pp. \bfpage{1126}--\blpage{1133}
(\byear{2019})
\end{bchapter}
\endbibitem

\bibitem[\protect\citeauthoryear{Mi et~al.}{2024}]{mi2024acdnet}
\begin{barticle}
\bauthor{\bsnm{Mi}, \binits{J.}},
\bauthor{\bsnm{Zu}, \binits{Y.}},
\bauthor{\bsnm{Wang}, \binits{Z.}},
\bauthor{\bsnm{He}, \binits{J.}}:
\batitle{Acdnet: Attention-guided collaborative decision network for effective
  medication recommendation}.
\bjtitle{Journal of Biomedical Informatics}
\bvolume{149},
\bfpage{104570}
(\byear{2024})
\end{barticle}
\endbibitem

\bibitem[\protect\citeauthoryear{Cai et~al.}{2016}]{cai2016real}
\begin{barticle}
\bauthor{\bsnm{Cai}, \binits{X.}},
\bauthor{\bsnm{Perez-Concha}, \binits{O.}},
\bauthor{\bsnm{Coiera}, \binits{E.}},
\bauthor{\bsnm{Martin-Sanchez}, \binits{F.}},
\bauthor{\bsnm{Day}, \binits{R.}},
\bauthor{\bsnm{Roffe}, \binits{D.}},
\bauthor{\bsnm{Gallego}, \binits{B.}}:
\batitle{Real-time prediction of mortality, readmission, and length of stay
  using electronic health record data}.
\bjtitle{Journal of the American Medical Informatics Association}
\bvolume{23}(\bissue{3}),
\bfpage{553}--\blpage{561}
(\byear{2016})
\end{barticle}
\endbibitem

\bibitem[\protect\citeauthoryear{Sottile et~al.}{2021}]{sottile2021real}
\begin{barticle}
\bauthor{\bsnm{Sottile}, \binits{P.D.}},
\bauthor{\bsnm{Albers}, \binits{D.}},
\bauthor{\bsnm{DeWitt}, \binits{P.E.}},
\bauthor{\bsnm{Russell}, \binits{S.}},
\bauthor{\bsnm{Stroh}, \binits{J.}},
\bauthor{\bsnm{Kao}, \binits{D.P.}},
\bauthor{\bsnm{Adrian}, \binits{B.}},
\bauthor{\bsnm{Levine}, \binits{M.E.}},
\bauthor{\bsnm{Mooney}, \binits{R.}},
\bauthor{\bsnm{Larchick}, \binits{L.}}, \betal:
\batitle{Real-time electronic health record mortality prediction during the
  covid-19 pandemic: a prospective cohort study}.
\bjtitle{Journal of the American Medical Informatics Association}
\bvolume{28}(\bissue{11}),
\bfpage{2354}--\blpage{2365}
(\byear{2021})
\end{barticle}
\endbibitem

\bibitem[\protect\citeauthoryear{Shen et~al.}{2018}]{shen2018utilization}
\begin{barticle}
\bauthor{\bsnm{Shen}, \binits{F.}},
\bauthor{\bsnm{Liu}, \binits{S.}},
\bauthor{\bsnm{Wang}, \binits{Y.}},
\bauthor{\bsnm{Wen}, \binits{A.}},
\bauthor{\bsnm{Wang}, \binits{L.}},
\bauthor{\bsnm{Liu}, \binits{H.}}, \betal:
\batitle{Utilization of electronic medical records and biomedical literature to
  support the diagnosis of rare diseases using data fusion and collaborative
  filtering approaches}.
\bjtitle{JMIR medical informatics}
\bvolume{6}(\bissue{4}),
\bfpage{11301}
(\byear{2018})
\end{barticle}
\endbibitem

\bibitem[\protect\citeauthoryear{Farhan et~al.}{2016}]{farhan2016predictive}
\begin{barticle}
\bauthor{\bsnm{Farhan}, \binits{W.}},
\bauthor{\bsnm{Wang}, \binits{Z.}},
\bauthor{\bsnm{Huang}, \binits{Y.}},
\bauthor{\bsnm{Wang}, \binits{S.}},
\bauthor{\bsnm{Wang}, \binits{F.}},
\bauthor{\bsnm{Jiang}, \binits{X.}}, \betal:
\batitle{A predictive model for medical events based on contextual embedding of
  temporal sequences}.
\bjtitle{JMIR medical informatics}
\bvolume{4}(\bissue{4}),
\bfpage{5977}
(\byear{2016})
\end{barticle}
\endbibitem

\bibitem[\protect\citeauthoryear{Su and Khoshgoftaar}{2009}]{su2009survey}
\begin{barticle}
\bauthor{\bsnm{Su}, \binits{X.}},
\bauthor{\bsnm{Khoshgoftaar}, \binits{T.M.}}:
\batitle{A survey of collaborative filtering techniques}.
\bjtitle{Advances in artificial intelligence}
\bvolume{2009}(\bissue{1}),
\bfpage{421425}
(\byear{2009})
\end{barticle}
\endbibitem

\bibitem[\protect\citeauthoryear{Yang et~al.}{2014}]{YANG20141}
\begin{barticle}
\bauthor{\bsnm{Yang}, \binits{X.}},
\bauthor{\bsnm{Guo}, \binits{Y.}},
\bauthor{\bsnm{Liu}, \binits{Y.}},
\bauthor{\bsnm{Steck}, \binits{H.}}:
\batitle{A survey of collaborative filtering based social recommender systems}.
\bjtitle{Computer Communications}
\bvolume{41},
\bfpage{1}--\blpage{10}
(\byear{2014})
\doiurl{10.1016/j.comcom.2013.06.009}
\end{barticle}
\endbibitem

\bibitem[\protect\citeauthoryear{Huang et~al.}{2023}]{HUANG2023200307}
\begin{barticle}
\bauthor{\bsnm{Huang}, \binits{Y.}},
\bauthor{\bsnm{Loux}, \binits{T.}},
\bauthor{\bsnm{Huang}, \binits{X.}},
\bauthor{\bsnm{Feng}, \binits{X.}}:
\batitle{The relationship between chronic diseases and mental health: A
  cross-sectional study}.
\bjtitle{Mental Health \& Prevention}
\bvolume{32},
\bfpage{200307}
(\byear{2023})
\doiurl{10.1016/j.mhp.2023.200307}
\end{barticle}
\endbibitem

\bibitem[\protect\citeauthoryear{Sun et~al.}{2014}]{sun2014predicting}
\begin{barticle}
\bauthor{\bsnm{Sun}, \binits{K.}},
\bauthor{\bsnm{Gon{\c{c}}alves}, \binits{J.P.}},
\bauthor{\bsnm{Larminie}, \binits{C.}},
\bauthor{\bsnm{Pr{\v{z}}ulj}, \binits{N.}}:
\batitle{Predicting disease associations via biological network analysis}.
\bjtitle{BMC bioinformatics}
\bvolume{15},
\bfpage{1}--\blpage{13}
(\byear{2014})
\end{barticle}
\endbibitem

\bibitem[\protect\citeauthoryear{Hao and Blair}{2016}]{hao2016comparative}
\begin{barticle}
\bauthor{\bsnm{Hao}, \binits{F.}},
\bauthor{\bsnm{Blair}, \binits{R.H.}}:
\batitle{A comparative study: classification vs. user-based collaborative
  filtering for clinical prediction}.
\bjtitle{BMC medical research methodology}
\bvolume{16},
\bfpage{1}--\blpage{14}
(\byear{2016})
\end{barticle}
\endbibitem

\bibitem[\protect\citeauthoryear{Sae-Ang et~al.}{2022}]{sae2022drug}
\begin{barticle}
\bauthor{\bsnm{Sae-Ang}, \binits{A.}},
\bauthor{\bsnm{Chairat}, \binits{S.}},
\bauthor{\bsnm{Tansuebchueasai}, \binits{N.}},
\bauthor{\bsnm{Fumaneeshoat}, \binits{O.}},
\bauthor{\bsnm{Ingviya}, \binits{T.}},
\bauthor{\bsnm{Chaichulee}, \binits{S.}}:
\batitle{Drug recommendation from diagnosis codes: classification vs.
  collaborative filtering approaches}.
\bjtitle{International Journal of Environmental Research and Public Health}
\bvolume{20}(\bissue{1}),
\bfpage{309}
(\byear{2022})
\end{barticle}
\endbibitem

\bibitem[\protect\citeauthoryear{Johnson et~al.}{2023a}]{mimiciv_v2}
\begin{botherref}
\oauthor{\bsnm{Johnson}, \binits{A.}},
\oauthor{\bsnm{Bulgarelli}, \binits{L.}},
\oauthor{\bsnm{Pollard}, \binits{T.}},
\oauthor{\bsnm{Horng}, \binits{S.}},
\oauthor{\bsnm{Celi}, \binits{L.A.}},
\oauthor{\bsnm{Mark}, \binits{R.}}:
MIMIC-IV.
PhysioNet
(2023).
\doiurl{10.13026/6mm1-ek67} .
\url{https://physionet.org/content/mimiciv/2.2/}
\end{botherref}
\endbibitem

\bibitem[\protect\citeauthoryear{Johnson et~al.}{2023b}]{johnson2023mimic}
\begin{barticle}
\bauthor{\bsnm{Johnson}, \binits{A.E.}},
\bauthor{\bsnm{Bulgarelli}, \binits{L.}},
\bauthor{\bsnm{Shen}, \binits{L.}},
\bauthor{\bsnm{Gayles}, \binits{A.}},
\bauthor{\bsnm{Shammout}, \binits{A.}},
\bauthor{\bsnm{Horng}, \binits{S.}},
\bauthor{\bsnm{Pollard}, \binits{T.J.}},
\bauthor{\bsnm{Hao}, \binits{S.}},
\bauthor{\bsnm{Moody}, \binits{B.}},
\bauthor{\bsnm{Gow}, \binits{B.}}, \betal:
\batitle{Mimic-iv, a freely accessible electronic health record dataset}.
\bjtitle{Scientific data}
\bvolume{10}(\bissue{1}),
\bfpage{1}
(\byear{2023})
\end{barticle}
\endbibitem

\bibitem[\protect\citeauthoryear{Maaten and
  Hinton}{2008}]{maaten2008visualizing}
\begin{barticle}
\bauthor{\bsnm{Maaten}, \binits{L.v.d.}},
\bauthor{\bsnm{Hinton}, \binits{G.}}:
\batitle{Visualizing data using t-sne}.
\bjtitle{Journal of machine learning research}
\bvolume{9}(\bissue{Nov}),
\bfpage{2579}--\blpage{2605}
(\byear{2008})
\end{barticle}
\endbibitem

\bibitem[\protect\citeauthoryear{for Medicare et~al.}{2017}]{centers2017icd}
\begin{barticle}
\bauthor{\bsnm{Medicare}, \binits{C.}},
\bauthor{\bsnm{(CMS)}, \binits{M.S.}}, \betal:
\batitle{Icd-10-cm general equivalence mappings (gems)--diagnosis codes}.
\bjtitle{Published online August}
\bvolume{22},
\bfpage{2016}
(\byear{2017})
\end{barticle}
\endbibitem

\bibitem[\protect\citeauthoryear{Sechi and Serra}{2007}]{sechi2007wernicke}
\begin{barticle}
\bauthor{\bsnm{Sechi}, \binits{G.}},
\bauthor{\bsnm{Serra}, \binits{A.}}:
\batitle{Wernicke's encephalopathy: new clinical settings and recent advances
  in diagnosis and management}.
\bjtitle{The Lancet Neurology}
\bvolume{6}(\bissue{5}),
\bfpage{442}--\blpage{455}
(\byear{2007})
\end{barticle}
\endbibitem

\bibitem[\protect\citeauthoryear{Libby et~al.}{2002}]{libby2002inflammation}
\begin{barticle}
\bauthor{\bsnm{Libby}, \binits{P.}},
\bauthor{\bsnm{Ridker}, \binits{P.M.}},
\bauthor{\bsnm{Maseri}, \binits{A.}}:
\batitle{Inflammation and atherosclerosis}.
\bjtitle{Circulation}
\bvolume{105}(\bissue{9}),
\bfpage{1135}--\blpage{1143}
(\byear{2002})
\end{barticle}
\endbibitem

\bibitem[\protect\citeauthoryear{Stone et~al.}{2014}]{stone20142013}
\begin{barticle}
\bauthor{\bsnm{Stone}, \binits{N.J.}},
\bauthor{\bsnm{Robinson}, \binits{J.G.}},
\bauthor{\bsnm{Lichtenstein}, \binits{A.H.}},
\bauthor{\bsnm{Bairey~Merz}, \binits{C.N.}},
\bauthor{\bsnm{Blum}, \binits{C.B.}},
\bauthor{\bsnm{Eckel}, \binits{R.H.}},
\bauthor{\bsnm{Goldberg}, \binits{A.C.}},
\bauthor{\bsnm{Gordon}, \binits{D.}},
\bauthor{\bsnm{Levy}, \binits{D.}},
\bauthor{\bsnm{Lloyd-Jones}, \binits{D.M.}}, \betal:
\batitle{2013 acc/aha guideline on the treatment of blood cholesterol to reduce
  atherosclerotic cardiovascular risk in adults: a report of the american
  college of cardiology/american heart association task force on practice
  guidelines}.
\bjtitle{Journal of the American College of Cardiology}
\bvolume{63}(\bissue{25 Part B}),
\bfpage{2889}--\blpage{2934}
(\byear{2014})
\end{barticle}
\endbibitem

\bibitem[\protect\citeauthoryear{Rendle et~al.}{2012}]{rendle2012bpr}
\begin{botherref}
\oauthor{\bsnm{Rendle}, \binits{S.}},
\oauthor{\bsnm{Freudenthaler}, \binits{C.}},
\oauthor{\bsnm{Gantner}, \binits{Z.}},
\oauthor{\bsnm{Schmidt-Thieme}, \binits{L.}}:
Bpr: Bayesian personalized ranking from implicit feedback.
arXiv preprint arXiv:1205.2618
(2012)
\end{botherref}
\endbibitem

\bibitem[\protect\citeauthoryear{He and
  Chua}{2017}]{he2017neuralfactorizationmachinessparse}
\begin{botherref}
\oauthor{\bsnm{He}, \binits{X.}},
\oauthor{\bsnm{Chua}, \binits{T.-S.}}:
Neural Factorization Machines for Sparse Predictive Analytics
(2017).
\url{https://arxiv.org/abs/1708.05027}
\end{botherref}
\endbibitem

\bibitem[\protect\citeauthoryear{Wang et~al.}{2019}]{wang2019neural}
\begin{bchapter}
\bauthor{\bsnm{Wang}, \binits{X.}},
\bauthor{\bsnm{He}, \binits{X.}},
\bauthor{\bsnm{Wang}, \binits{M.}},
\bauthor{\bsnm{Feng}, \binits{F.}},
\bauthor{\bsnm{Chua}, \binits{T.-S.}}:
\bctitle{Neural graph collaborative filtering}.
In: \bbtitle{Proceedings of the 42nd International ACM SIGIR Conference on
  Research and Development in Information Retrieval},
pp. \bfpage{165}--\blpage{174}
(\byear{2019})
\end{bchapter}
\endbibitem

\bibitem[\protect\citeauthoryear{He
  et~al.}{2020}]{he2020lightgcnsimplifyingpoweringgraph}
\begin{botherref}
\oauthor{\bsnm{He}, \binits{X.}},
\oauthor{\bsnm{Deng}, \binits{K.}},
\oauthor{\bsnm{Wang}, \binits{X.}},
\oauthor{\bsnm{Li}, \binits{Y.}},
\oauthor{\bsnm{Zhang}, \binits{Y.}},
\oauthor{\bsnm{Wang}, \binits{M.}}:
LightGCN: Simplifying and Powering Graph Convolution Network for Recommendation
(2020).
\url{https://arxiv.org/abs/2002.02126}
\end{botherref}
\endbibitem

\bibitem[\protect\citeauthoryear{Wang et~al.}{2019}]{wang2019kgat}
\begin{bchapter}
\bauthor{\bsnm{Wang}, \binits{X.}},
\bauthor{\bsnm{He}, \binits{X.}},
\bauthor{\bsnm{Cao}, \binits{Y.}},
\bauthor{\bsnm{Liu}, \binits{M.}},
\bauthor{\bsnm{Chua}, \binits{T.-S.}}:
\bctitle{Kgat: Knowledge graph attention network for recommendation}.
In: \bbtitle{Proceedings of the 25th ACM SIGKDD International Conference on
  Knowledge Discovery \& Data Mining},
pp. \bfpage{950}--\blpage{958}
(\byear{2019})
\end{bchapter}
\endbibitem

\bibitem[\protect\citeauthoryear{Yang et~al.}{2023}]{yang2023manydg}
\begin{botherref}
\oauthor{\bsnm{Yang}, \binits{C.}},
\oauthor{\bsnm{Westover}, \binits{M.B.}},
\oauthor{\bsnm{Sun}, \binits{J.}}:
Manydg: Many-domain generalization for healthcare applications.
arXiv preprint arXiv:2301.08834
(2023)
\end{botherref}
\endbibitem

\bibitem[\protect\citeauthoryear{Kipf and
  Welling}{2017}]{kipf2017semisupervised}
\begin{bchapter}
\bauthor{\bsnm{Kipf}, \binits{T.N.}},
\bauthor{\bsnm{Welling}, \binits{M.}}:
\bctitle{Semi-supervised classification with graph convolutional networks}.
In: \bbtitle{International Conference on Learning Representations}
(\byear{2017}).
\burl{https://openreview.net/forum?id=SJU4ayYgl}
\end{bchapter}
\endbibitem

\bibitem[\protect\citeauthoryear{Maas et~al.}{2013}]{maas2013rectifier}
\begin{bchapter}
\bauthor{\bsnm{Maas}, \binits{A.L.}},
\bauthor{\bsnm{Hannun}, \binits{A.Y.}},
\bauthor{\bsnm{Ng}, \binits{A.Y.}}, \betal:
\bctitle{Rectifier nonlinearities improve neural network acoustic models}.
In: \bbtitle{Proc. Icml},
vol. \bseriesno{30},
p. \bfpage{3}
(\byear{2013}).
\bcomment{Atlanta, GA}
\end{bchapter}
\endbibitem

\bibitem[\protect\citeauthoryear{He et~al.}{2017}]{he2017neural}
\begin{bchapter}
\bauthor{\bsnm{He}, \binits{X.}},
\bauthor{\bsnm{Liao}, \binits{L.}},
\bauthor{\bsnm{Zhang}, \binits{H.}},
\bauthor{\bsnm{Nie}, \binits{L.}},
\bauthor{\bsnm{Hu}, \binits{X.}},
\bauthor{\bsnm{Chua}, \binits{T.-S.}}:
\bctitle{Neural collaborative filtering}.
In: \bbtitle{Proceedings of the 26th International Conference on World Wide
  Web},
pp. \bfpage{173}--\blpage{182}
(\byear{2017})
\end{bchapter}
\endbibitem

\end{thebibliography}

\section*{Author contributions}
\noindent H.Z. and Y.B. are joint first authors. J.F. proposed the original ideas of the work, and J.C. improved the ideas. H.Z. preprocessed the data and wrote the code of the algorithms. H.Z., Y.B., Q.X., and L.Z. wrote the draft of the paper. Q.X. and L.Z. contributed to the visualization, and Y.B. analyzed the results. All authors had access to the data and have read and approved the final manuscript. The corresponding author attests that all listed authors meet authorship criteria and that no others meeting the criteria have been omitted. J.C. and J.F. are the guarantors.

\section*{Competing interests}
\noindent The authors declare no competing interests.

\section*{Additional information}
\noindent \textbf{Supplementary information} Supplementary Information is available for this paper at supplementary.pdf. \\

\noindent \textbf{Correspondence} and requests for materials should be addressed to Jicong Fan.

\end{document}